\documentclass[sigconf]{acmart}

\usepackage{enumitem}
\usepackage{mathtools}
\definecolor{darkred}{RGB}{162, 0, 0}
\newcommand{\darkred}[1]{\textbf{\textcolor{darkred}{#1}}}
\usepackage{multirow}
\acmSubmissionID{262}

\AtBeginDocument{%
  \providecommand\BibTeX{{%
    \normalfont B\kern-0.5em{\scshape i\kern-0.25em b}\kern-0.8em\TeX}}}

\copyrightyear{2022}
\acmYear{2022}
\setcopyright{acmcopyright}
\acmConference[WSDM '22]{Proceedings of the Fifteenth ACM International Conference on Web Search and Data Mining}{February 21--25, 2022}{Tempe, AZ, USA}
\acmBooktitle{Proceedings of the Fifteenth ACM International Conference on Web Search and Data Mining (WSDM '22), February 21--25, 2022, Tempe, AZ, USA}
\acmPrice{15.00}
\acmDOI{10.1145/3488560.3498416}
\acmISBN{978-1-4503-9132-0/22/02}

\begin{document}
\fancyhead{}

\title{Bringing Your Own View: Graph Contrastive Learning without Prefabricated Data Augmentations}


\author{Yuning You}
\affiliation{%
  \institution{Texas A\&M University}
  \streetaddress{}
  \country{}
  }
\email{yuning.you@tamu.edu}

\author{Tianlong Chen}
\affiliation{%
  \institution{University of Texas at Austin}
  \streetaddress{}
  \country{}
  }
\email{tianlong.chen@utexas.edu}

\author{Zhangyang Wang}
\affiliation{%
  \institution{University of Texas at Austin}
  \streetaddress{}
  \country{}
  }
\email{atlaswang@utexas.edu}

\author{Yang Shen}
\affiliation{%
  \institution{Texas A\&M University}
  \streetaddress{}
  \country{}
  }
\email{yshen@tamu.edu}

\renewcommand{\shortauthors}{Trovato and Tobin, et al.}

\begin{abstract}
  Self-supervision is recently surging at its new frontier of graph learning.  It facilitates graph representations beneficial to downstream tasks; but its success could hinge on domain knowledge for handcraft or the often expensive trials and errors.  Even its state-of-the-art representative, graph contrastive learning (GraphCL), is not completely free of those needs as GraphCL uses a prefabricated prior reflected by the ad-hoc manual selection of graph data augmentations. Our work aims at advancing GraphCL by answering the following questions: \textit{How to represent the space of graph augmented views? What principle can be relied upon to learn a prior in that space? And what framework can be constructed to learn the prior in tandem with contrastive learning?} Accordingly, we have extended the prefabricated discrete prior in the augmentation set, to a learnable continuous prior in the parameter space of graph generators, assuming that graph priors \textit{per se}, similar to the concept of image manifolds, can be learned by data generation. Furthermore, to form contrastive views without collapsing to trivial solutions due to the prior learnability, we have leveraged both principles of information minimization (InfoMin) and information bottleneck (InfoBN) to regularize the learned priors. Eventually, contrastive learning, InfoMin, and InfoBN are incorporated organically into one framework of bi-level optimization. Our principled and automated approach has proven to be competitive against the state-of-the-art graph self-supervision methods, including GraphCL, on benchmarks of small graphs; and shown even better generalizability on large-scale graphs, without resorting to human expertise or downstream validation.
  Our code is publicly released at \url{https: //github.com/Shen-Lab/GraphCL_Automated}.
\end{abstract}




\maketitle

\section{Introduction}
Self-supervised learning on non-Euclidean structured data  has recently intrigued vast interest, with the capability of learning generalizable, transferable and robust representations from unlabeled graph data \cite{xie2021self,liu2021graph,hu2019strategies,you2020graph,liu2020self}.
Unlike images, speeches or natural languages, graph-structured data are not monomorphic but abstractions of \textit{diverse nature} (e.g. social networks, polymers or power grids \cite{kipf2016semi,zou2019layer,you2020cross,you2020l2}).
This unique heterogeneity challenge however has not been fully addressed in the previous self-supervised works.
The success of existing approaches relies on carefully designed predictive pretext tasks with domain expertise (e.g. context prediction \cite{hu2019strategies}, meta-path extraction \cite{hwang2020self}, graph completion \cite{you2020does}, etc
\cite{perozzi2014deepwalk,tang2015line,kipf2016variational,hamilton2017inductive,hu2020gpt,sun2020multi,sehanobish2020self,jin2020self,rong2020self,hao2021pre,kim2021find,yu2021self,zhang2020iterative,hwang2021self,li2021representation,huang2021hop,manessi2021graph,suresh2021adversarial,xu2021infogcl,kefatoself,xu2021group}),
possessing the premise that the designated task is the generally effective \textit{prior} across all datasets, while it is not always guaranteed especially in the mentioned diversity context.
The recently emerged contrastive methods seem to be free from setting the pretext, which whilst exists in a disguised form:
appropriate hand-crafted contrastive views are required to be constructed (e.g. global-local representations \cite{sun2019infograph}, diffusion matrices \cite{hassani2020contrastive}, $r$-ego networks \cite{qiu2020gcc}, etc
\cite{velivckovic2018deep,peng2020self,zhu2020graph,chen2020distance,chen2020coad,ren2019heterogeneous,park2020unsupervised,peng2020graph,zhang2020motif,roy2021node})
otherwise resulting in performance degrade \cite{sun2019infograph,you2020graph}.
The state-of-the-art (SOTA) representative, graph contrastive learning (GraphCL) \cite{you2020graph} even copes with this challenge with additional human labors: it contrasts on the augmented graphs via manually selecting and applying the prefabricated augmentation operations per dataset \cite{you2020graph,zhao2020data,kong2020flag,verma2019graphmix,jin2020self,jin2021automated,you2021graph}, by either rule of thumb or trial-and-errors.
Thus, it is more flexible in accordance with diverse graph datasets,
although more expensive since the rules are derived by tedious tuning with downstream labels and built on top of a pool of prefabricated priors, i.e. off-the-shelf augmentations.

Our perspective to help close the gap is to turn the prefabricated self-supervised prior into a \textit{learnable} one.
Intuitively, the learnable prior,  following the data-driven ethos, is more versatile compared with sticking to the unaltered one and less resource-demanding compared with manual picking from ready-made ones.
Learned priors were explored in video generation, compressed sensing and Bayesian deep learning
\cite{denton2018stochastic,bora2017compressed,tripathi2018correction,grover2018amortized,kabkab2018task,shah2018solving,fletcher2018inference,asim2018solving,hand2018global,fortuin2021priors,ulyanov2018deep}.
To the best of our knowledge, this perspective however has not been explored in the discrete and irregular data structure of graphs.

\begin{figure*}[t]
  \begin{center}
    \includegraphics[width=0.75\textwidth]{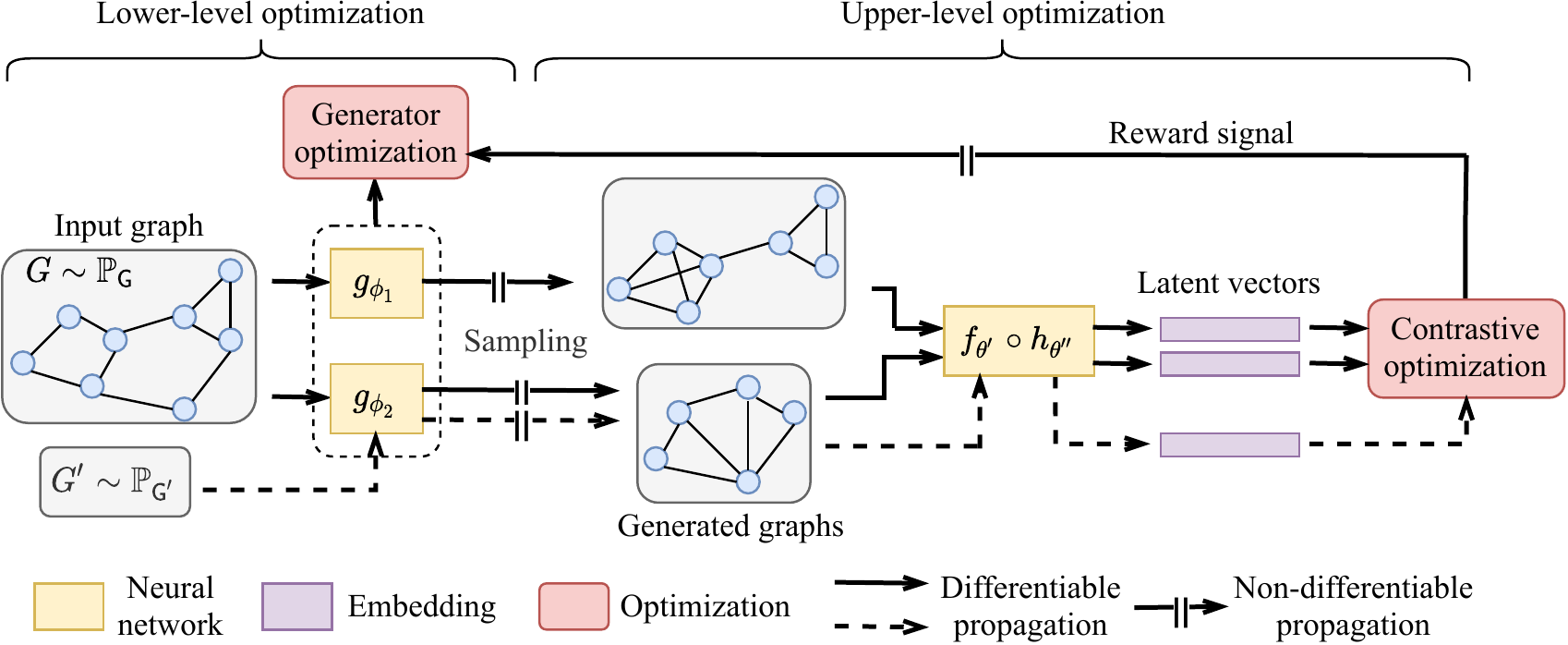}
  \end{center}
  \caption{Pipeline of GraphCL with learned prior. Graph generative models $g_{\phi_1}, g_{\phi_2}$ generate contrastive views for self-supervised contrasting, and then receive the reward for their parameter update.} \label{fig:graphcl_lp}
\end{figure*}

\textbf{Contributions.}
In this paper, we aim at tackling the aforementioned challenge by asking the following fundamental question: \textit{What is the space, principle and framework that one can rely on, to define and pursue the learnable    self-supervised prior?}
We offer the following answers accordingly.
Leveraging the SOTA GraphCL framework as the base model,
we innovate in extending the prefabricated discrete prior in the augmentation set into a learnable continuous prior parameterized by a neural network, that adaptively and dynamically learns from data during contrastive training.
(i) The prior \textit{space} is determined by the parameter space of the neural network, for which we utilize graph generative models  
\cite{chakrabarti2006graph,kipf2016variational,wang2018graphgan,bojchevski2018netgan,shi2020graphaf,liu2021graphebm,luo2021graphdf,jin2018junction,you2018graphrnn,luo2021graphdf}
for the parameterization.
The assumption we make here is that, similar to images, the graph prior can also be well-captured by data generation
\cite{denton2018stochastic,bora2017compressed,tripathi2018correction,grover2018amortized,kabkab2018task,shah2018solving,fletcher2018inference,asim2018solving,hand2018global,fortuin2021priors}
(ii) Besides, to avoid collapsing to the trivial solution \cite{bardes2021vicreg,chen2020simple}, we employ the information minimization (InfoMin) and information bottleneck (InfoBN) as  \textit{principles} \cite{tian2020makes,tishby2000information,alemi2016deep} to regularize the generator optimization.
(iii) Ultimately, the new approach is mathematically formulated as a bi-level optimization (\textit{framework}) (see Figure \ref{fig:graphcl_lp} and optimization \eqref{eq:graphcl_learned_prior_reward}), with the proposed components briefly summarized as follows:

\begin{itemize}[leftmargin=*]
  \item Learnable prior function parameterized by a graph generative model, which is presumably capable of  well-capturing the graph prior from data (see Section \ref{sec:learned_prior}); and 
  \item Principles of InfoMin and InfoBN to regularize generator optimization during contrastive learning, avoiding the trivial solution of collapses (see Section \ref{sec:principle}).
\end{itemize}

Although GraphCL is used as the base model in this paper, we need to emphasize that the proposed learned prior is flexible enough and easy to be adopted in other contrastive methods.

We hereby highlight our contributions, that (i) we make the first attempt to incorporate the learnable prior with graph neural network, further exploiting the power of abundant data under the instructive precedent assumption, and
(ii) we learn in adaptivity and automation, that not only requires little human effort to prefabricate priors, i.e. augmentation functions, but also learns such knowledge during self-supervision in a data-driven, flexible and principled fashion.
This is crucial for better generalizability in accordance with the graph polymorphism challenge, which is achieved in a data-driven fashion without handcrafted knowledge or expensive trial-and-error, especially on datasets of large scale.

We evaluate our proposed methods in semi-supervised learning and transfer learning \cite{you2020graph,hu2019strategies} on graph datasets including social networks, protein-protein interaction networks, code abstract syntax trees and molecules \cite{Morris+2020,hu2020open,hu2019strategies}.
Numerically, we show that the learned prior in GraphCL performs on par with the SOTA competitors on small benchmarks, and generalizes better on large-scale datasets, without resorting to any human labor on pre-defining augmentations or  tedious tuning.

\section{Related Work}
\begin{figure}[!htb]
  \begin{center}
    \includegraphics[width=0.38\textwidth]{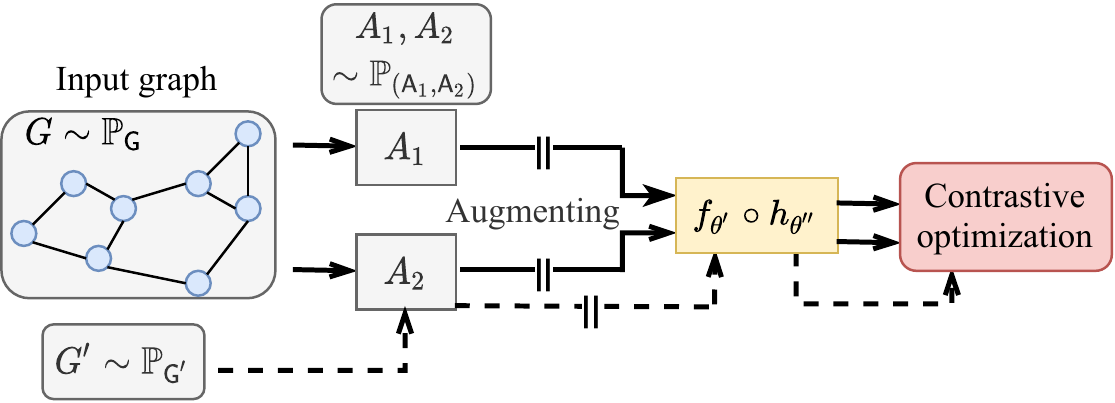}
  \end{center} \vspace{-1.5em}
  \caption{Pipeline of Graph Contrastive Learning with augmentations (GraphCL). 
  } \label{fig:graphcl} \vspace{-1.5em}
\end{figure}

\textbf{Graph contrastive learning.}
Contrastive learning on graph data is shown to be a promising technique in graph representation learning via exploiting abundant unlabelled data
\cite{velivckovic2018deep,sun2019infograph,you2020graph}.
The state-of-the-art graph contrastive learning framework (GraphCL) \cite{you2020graph} emphasizes the perturbation invariance in graph neural networks (GNNs) through maximizing agreement between two augmented graph views, with an  overview in Figure \ref{fig:graphcl}.

Mathematically\footnote{We use the sans-serif typeface to denote a random variable (e.g. $\mathsf{G}$). The same letter in italics (e.g. $G$) denotes a sample and that in the calligraphic font (e.g. $\mathcal{G}$) denotes the sample space.},
we denote the graph-structured input $G \in \mathcal{G}$ sampled from certain empirical distribution $\mathbb{P}_{\mathsf{G}}$, that $G = \{V, E\}$ is an undirected graph with $V$, $E$ denoting the set of nodes and edges, and $X_v \in \mathcal{R}^{D_V}$ for $v \in V$, $X_e \in \mathcal{R}^{D_E}$ for $e \in E$ denoting node and edge features, respectively.
GraphCL samples two augmentation operators $A_1, A_2 \sim \mathbb{P}_{\mathsf{A}_1, \mathsf{A}_2}$ from a given augmentation pool $\mathcal{A} =  \{ \mathrm{NodeDrop}, \mathrm{Subgraph}, \mathrm{EdgePert}, \mathrm{AttrMask}, \mathrm{Identical} \}$ \cite{you2020graph,zhao2020data,kong2020flag,verma2019graphmix} that $A \in \mathcal{A}: \mathcal{G} \rightarrow \mathcal{G}$ is the stochastic augmentation function \cite{stochastic_function},
and optimizes the following contrastive loss:
\begin{align} \label{eq:graphcl}
    & \min_\theta \; \mathbb{E}_{\mathbb{P}_{\mathsf{G}}} \mathcal{L}_{\mathrm{CL}}(\mathsf{G}, \mathsf{A}_1, \mathsf{A}_2, \theta) \notag \\
    =\, & \min_\theta \; \mathbb{E}_{\mathbb{P}_{\mathsf{G}}} \bigg\{ - \mathbb{E}_{\mathbb{P}_{(\mathsf{A}_1, \mathsf{A}_2)}
    } \mathrm{sim}\big( {\overbrace{\textstyle \mathsf{T}_{\theta, 1}(\mathsf{G}), \mathsf{T}_{\theta, 2}(\mathsf{G})}^{\mathclap{\text{positive pairs}}} } \big) \notag \\
    & + \mathbb{E}_{\mathbb{P}_{\mathsf{A}_1}} \mathrm{log}\bigg(\mathbb{E}_{\mathbb{P}_{\mathsf{G}'} \times \mathbb{P}_{\mathsf{A}_2}} \mathrm{exp}\Big(\mathrm{sim}\big( {\underbrace{\mathsf{T}_{\theta, 1}(\mathsf{G}), \mathsf{T}_{\theta, 2}(\mathsf{G}')}_{\mathclap{\text{negative pairs}}} }\big)\Big)\bigg)\bigg\},
\end{align}
where $\mathsf{T}_{\theta, i} = \mathsf{A}_i \circ f_{\theta'} \circ h_{\theta''} \, (i=1,2)$ is parameterized by $\theta = \{ \theta', \theta'' \}$,
$f_{\theta'}: \mathcal{G} \rightarrow \mathcal{R}^{D'}$ is the GNN to be pre-trained,
$h_{\theta''}: \mathcal{R}^{D'} \rightarrow \mathcal{R}^{D''}$ is the projection head,
$\mathrm{sim}(u, v) = \frac{u^\mathsf{T}v}{\lVert u \rVert \lVert v \rVert}$ is the cosine similarity function,
$\mathbb{P}_{\mathsf{G}'} = \mathbb{P}_{\mathsf{G}}$ is the negative sampling distribution,
and $\mathbb{P}_{\mathsf{A}_1}$ and $\mathbb{P}_{\mathsf{A}_2}$ are the marginal distributions.
Here $\mathbb{P}_{(\mathsf{A}_1, \mathsf{A}_2)}$ is heuristically pre-defined per dataset \cite{you2020graph}.
After contrastive pre-training, the pre-trained $f_{\theta'^*}$ can be further leveraged for downstream fine-tuning.

\textbf{Learnable prior.}
The learned priors were explored in video generation, compressed sensing (CS) and Bayesian deep learning 
\cite{denton2018stochastic,bora2017compressed,tripathi2018correction}.
SVG-LP \cite{denton2018stochastic} utilizes a recurrent neural network to model the temporally dependent priors, interpreted as an uncertainty predictor in generating video frames;
a branch of approaches in CS
\cite{bora2017compressed,tripathi2018correction,grover2018amortized}
shows that deep generative models can be used as extraordinary priors for images, with significantly fewer measurements compared to Lasso for a given reconstruction error;
and there are plenty of prior designs for Gaussian processes, variational autoencoders and Bayesian neural networks \cite{fortuin2021priors}.
Nevertheless, it has not been explored for graph data yet.

\textbf{Graph generative model.}
Generative models for graph data owns a longstanding history, with applications including molecular generation, anomaly detection and recommendation \cite{chakrabarti2006graph}.
We here focus on recent learning-based generative models  \cite{kipf2016variational,wang2018graphgan,bojchevski2018netgan}
conditioning on the input graph $G$, that the $\phi$-parameterized stochastic generation function is defined as $g_\phi: \mathcal{G} \rightarrow \mathcal{G}$ with the objective $\mathcal{L}_{\mathrm{Gen}}(\mathsf{G}, \phi)$ to optimize.

\section{Methodology} \label{sec:method}
\subsection{Graph Generative Model as Learnable Priors} \label{sec:learned_prior}
\textbf{Formulating the prior in GraphCL.}
We start with illustrating how we define the prior in our problem before detailing the proposed techniques.
We treat the prior in graph self-supervised learning as \textit{the inductive bias in belief beneficial to the downstream performance} \cite{hu2019strategies,you2020graph}, which we aim to enforce in our model via self-supervision.
Thus, such a prior is usually incorporated in the pretext tasks of the predictive methods, or reflected in the constructed augmentations/views of the contrastive-learning methods.
In GraphCL \cite{you2020graph}, it is reflected in the choice of the graph augmentation operators \cite{you2020graph,zhao2020data,kong2020flag,verma2019graphmix}.  For instance graph augmentation using node dropping encodes the prior that missing a random vertex does not alter downstream semantics.

Following this line of thoughts, we formally define the prior in GraphCL, as the stochastic mapping  \cite{stochastic_function} $m$ between graph manifolds that $m: \mathcal{G} \rightarrow \mathcal{G}$.
Notice that this definition is flexible that can be easily extended to other contrastive-learning methods (e.g. concatenating $m$ with another mapping $m': \mathcal{G} \rightarrow \mathcal{R}^D$ for methods demanding contrastive views in the vector space \cite{velivckovic2018deep,sun2019infograph}).

For clarification, the learned prior in GraphCL is different from that in Bayesian deep learning \cite{fortuin2021priors}, since it is defined for the irregular and discrete space of the graph manifold rather than the regular Euclidean space.
On the other hand, it is similar to a branch of works of the learned prior in compressed sensing leveraging generative models  
\cite{bora2017compressed,tripathi2018correction,grover2018amortized}, with the belief that the inductive bias encoded in the architecture/parameters provides beneficial prior knowledge.

Based on the above definition of priors in GraphCL, we can interpret its choice of augmentation functions as a kind of prior selection \cite{giannone2015prior,kass1996selection}.
Selecting a proper prior (for a specific dataset) was crucial for the performance of previous graph-learning methods \cite{hu2019strategies,you2020graph}. However, current prior selection was either burdened with tedious trial-and-error or restricted to a pool of prefabricated views, which limits its applicability and potential.

\textbf{A framework to incorporate a graph generative model-learned prior in contrastive learning.}
To close the aforementioned gaps in graph contrastive learning, we first propose the novel extension from the prefabricated discrete prior to a learnable continuous prior parameterized by a neural network (specifically, a graph generative model). Furthermore, the learned prior adaptively and dynamically evolves during contrastive training, which is detailed in the reward.

The recent rise of learning-based graph generative models \cite{kipf2016variational,wang2018graphgan,bojchevski2018netgan}
offers a smooth solution to parameterize a graph prior, where such a $\phi$-parameterized graph generator $g_\phi$, well-capturing graph distributions, is already a well-defined stochastic function between graph manifolds.  In this study we choose the widely-used variational graph auto-encoder (VGAE) \cite{kipf2016variational} as the generative model with the random-walk sampler \cite{wang2018graphgan}.
In addition to VGAE, other generators can also conveniently be the plug-and-play component in our framework \eqref{eq:graphcl_learned_prior_reward}.

A straightforward bi-level optimization form for GraphCL with the learned prior is thus written as:
\begin{align} \label{eq:graphcl_learned_prior}
    & \min_\theta \; \mathbb{E}_{\mathbb{P}_{\mathsf{G}}} \mathcal{L}_{\mathrm{CL}}(\mathsf{G}, \phi_1, \phi_2, \theta), \notag \\
    & \text{s.t.} \; \phi_1, \phi_2 \in \arg\min_{\phi_1', \phi_2'} \; \mathbb{E}_{\mathbb{P}_{\mathsf{G}}} \Big\{ \mathcal{L}_{\mathrm{Gen}}(\mathsf{G}, \phi_1') + \mathcal{L}_{\mathrm{Gen}}(\mathsf{G}, \phi_2') \Big\},
\end{align}
where the upper-level objective is for contrastive training and the lower-level constraint is for the generator optimization.  Compared to the formulation \eqref{eq:graphcl} with prefabricated graph views $\mathsf{A}$, the upper-level objective is now re-written with graph views from  $g_{\phi}$, that is, $\mathcal{L}_{\mathrm{CL}}(\mathsf{G}, g_{\phi_1}, g_{\phi_2}, \theta) = - \; \mathrm{sim}( \mathsf{T}_{\theta, \phi_1}(\mathsf{G}), \mathsf{T}_{\theta, \phi_2}(\mathsf{G}) ) + \mathrm{log}(\mathbb{E}_{\mathbb{P}_{\mathsf{G}'} } \mathrm{exp}(\mathrm{sim}($ $\mathsf{T}_{\theta, \phi_1}(\mathsf{G}), \mathsf{T}_{\theta, \phi_2}(\mathsf{G}') )))$, where $\mathsf{T}_{\theta, \phi_i} = g_{\phi_i} \circ f_{\theta'} \circ h_{\theta''} \, (i=1,2)$ and  $\theta = \{ \theta', \theta'' \}$. 

\textbf{Reward signal for graph generator.}
The straightforward formulation \eqref{eq:graphcl_learned_prior}, although sensible, has no message passing from the upper- to lower-level optimization during training. In other words, the generative model is trained in \eqref{eq:graphcl_learned_prior} regardless of GraphCL, which makes the prior learning non-adaptive to contrastive learning and could lead to trivial solutions of priors and mode collapse of contrastive learning (e.g. two graph generators output the same distribution, resulting in an easy contrastive optimization \cite{bardes2021vicreg,chen2020simple}).
In order to propagate feedback to the lower-level generator optimization, we additionally give it a ``reward'' signal (as depicted in Figure \ref{fig:graphcl_lp}) and reach a new formulation of bi-level optimization as 
\begin{align} \label{eq:graphcl_learned_prior_reward}
     \min_\theta \; & \; \mathbb{E}_{\mathbb{P}_{\mathsf{G}}} \mathcal{L}_{\mathrm{CL}}(\mathsf{G}, \phi_1, \phi_2, \theta), \notag \\
     \text{s.t.} \; & \; \phi_1, \phi_2 \in \arg\min_{\phi_1', \phi_2'} \; \mathbb{E}_{\mathbb{P}_{\mathsf{G}}} r(\mathsf{G}, \phi_1', \phi_2', \theta) \Big\{ \mathcal{L}_{\mathrm{Gen}}(\mathsf{G}, \phi_1') + \mathcal{L}_{\mathrm{Gen}}(\mathsf{G}, \phi_2') \Big\},
\end{align}
where the reward is in the simple form of $ r(\mathsf{G}, \phi_1, \phi_2, \theta) =$ \\ $\Big\{ \begin{smallmatrix} 1, & \text{given some condition} \\ \delta \ll 1, & \text{otherwise} \end{smallmatrix} $, with the condition determined by certain principles and the reward weakens to $\delta$ if the condition is not satisfied.
Optimization \eqref{eq:graphcl_learned_prior_reward} can be numerically solved by alternating gradient descent \cite{wang2019towards,boyd2004convex}.
The sensitivity analysis on $\delta$ is conducted in Section \ref{sec:hp_sensitivity}.

We will discuss the specific instantiation of the principles next in Section \ref{sec:principle}.
Without bells and whistles,  preliminary results in Table~\ref{tab:principles} demonstrate the necessity of the principled reward for better performance.

\begin{table}[!htb]
\begin{center}
 \caption{Preliminary experiments on learned prior (LP) w/ and w/o a principle on ogbg datasets. The principle is selected from \{InfoMin, InfoBN\} based on validation. \darkred{Red} fonts indicate  the best performances.} \label{tab:principles}
\resizebox{0.35\textwidth}{!}{\begin{tabular}{c c c c c c c c}
    \hline
  \hline
  Methods & ogbg-ppa & ogbg-code \\
  \hline
  \hline
  GraphCL & 57.77$\pm$1.25 & 22.45$\pm$0.17 \\
  \hline
  LP w/o principle & 58.13$\pm$1.80 & 23.16$\pm$0.24 \\
  LP w/ proper principle & \darkred{59.10}$\pm$0.88 & \darkred{23.50}$\pm$0.22 \\
  \hline
  \hline
\end{tabular}}
\end{center}
\end{table}

\subsection{Principles for Learning Priors to Contrast} \label{sec:principle}
\begin{figure}[!htb]
  \begin{center}
    \includegraphics[width=0.4\textwidth]{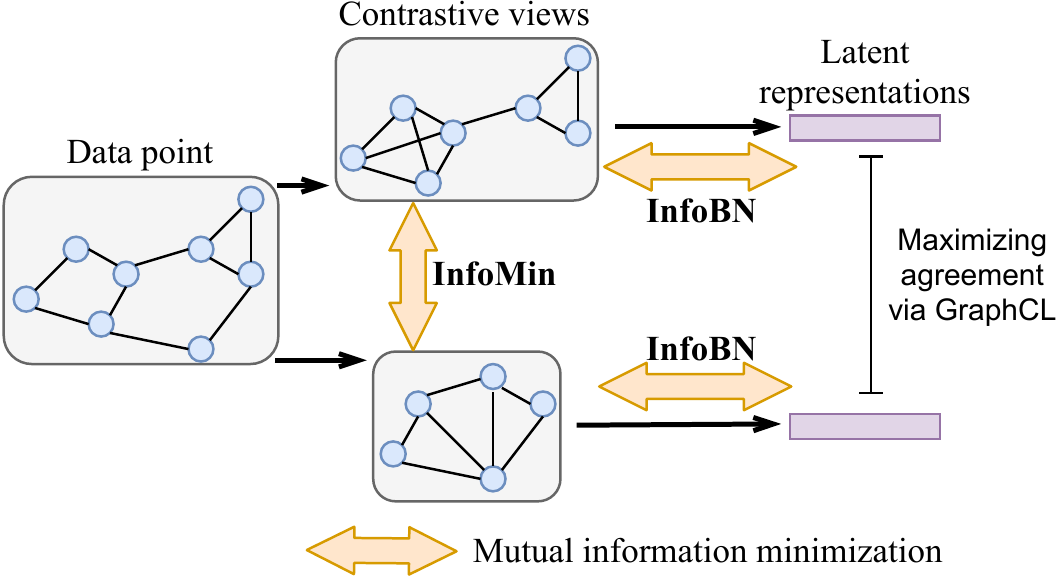}
  \end{center}
  \caption{Schematic diagram of the InfoMin and InfoBN principles to guide the prior learning in GraphCL.} \label{fig:principles}
\end{figure}

In this section we propose to adopt the principles of InfoMin and InfoBN and to incorporate them either or both, which guides the generator optimization to avoid collapsing to trivial solutions as discussed earlier. A schematic illustration is in Figure \ref{fig:principles}.

\textbf{Information minimization.}
The principle of information minimization (InfoMin)  \cite{tian2020makes,wang2019towards} encourages contrastive views to share less mutual information (MI) when maximizing the agreement between their latent representations. 
InfoMin thus can explicitly push generators to behave differently and  avoid collapse. Intuitively the MI minimization can make the encoders throw away more irrelevant information (or nuisance factors) to facilitate downstream generalization. 
Since the (negative) contrastive loss \eqref{eq:graphcl} is commonly adopted to act as a numerical MI estimator of the Donsker--Varadhan representation between views \cite{donsker1975asymptotic,belghazi2018mutual,tian2020makes,you2020graph},
we define the InfoMin reward function based on the estimated MI as $r_{\text{InfoMin}}(\mathsf{G}, \phi_1,$ $\phi_2, \theta) = \Big\{ \begin{smallmatrix} 1, & \text{if} \  \mathcal{L}_{\mathrm{CL}}(\mathsf{G}, \phi_1, \phi_2, \theta) > \text{threshold} \\ \delta \ll 1, & \text{otherwise} \end{smallmatrix}$,
where the threshold is treated as a hyper-parameter with sensitivity analysis in Section \ref{sec:hp_sensitivity}.

\textbf{Information bottleneck.}
The principle of information bottleneck (InfoBN)   \cite{tishby2000information,alemi2016deep,wu2020graph,yu2020graph} is originally proposed to discourage the representations from acquiring superfluous information that is irrelevant for predicting the target, aiming at better generalizability and robustness.
In the GraphCL framework, we introduce InfoBN by diminishing the information overlap between each contrastive view and its latent representation, whereas the pair of contrastive views still maintain a certain level of agreement in the latent space (through the upper-level contrastive loss).  
Thereby, the InfoBN reward is expressed as $r_{\text{InfoBN}}(\mathsf{G}, \phi_1, \phi_2, \theta, \pi) = \Big\{ \begin{smallmatrix} 1, & \text{if} \; \mathcal{L}_{\mathrm{InfoBN}}(\mathsf{G}, \phi_1, \phi_2, \theta, \pi) > \text{threshold} \\ \delta \ll 1, & \text{otherwise} \end{smallmatrix}$
where an additional MI estimator  between views and embeddings \cite{belghazi2018mutual,yu2020graph} is formulated as:
\begin{align}
    & \mathcal{L}_{\mathrm{InfoBN}}(\mathsf{G}, \phi_1, \phi_2, \theta, \pi) = \notag \\
    & - \mathrm{sim}( \mathsf{T}_{\pi, \phi_1}(\mathsf{G}), \mathsf{T}_{\theta, \phi_1}(\mathsf{G}) ) + \mathrm{log}(\mathbb{E}_{\mathbb{P}_{\mathsf{G'}} } \mathrm{exp}(\mathrm{sim}( \mathsf{T}_{\pi, \phi_1}(\mathsf{G}), \mathsf{T}_{\theta, \phi_1}(\mathsf{G'}) ))) \notag \\
    & - \mathrm{sim}( \mathsf{T}_{\pi, \phi_2}(\mathsf{G}), \mathsf{T}_{\theta, \phi_2}(\mathsf{G}) ) + \mathrm{log}(\mathbb{E}_{\mathbb{P}_{\mathsf{G'}} } \mathrm{exp}(\mathrm{sim}( \mathsf{T}_{\pi, \phi_2}(\mathsf{G}), \mathsf{T}_{\theta, \phi_2}(\mathsf{G'}) ))),
\end{align}
which needed to be minimized w.r.t. $\pi$ to achieve a precise estimation.
In this way, GraphCL with the InfoBN-rewarded learned prior is written as:
\begin{align} \label{eq:graphcl_learned_prior_reward_infobn}
    & \min_\theta \; \mathbb{E}_{\mathbb{P}_{\mathsf{G}}} \mathcal{L}_{\mathrm{CL}}(\mathsf{G}, \phi_1, \phi_2, \theta), \notag \\
    & \text{s.t.} \; \phi_1, \phi_2 \in \arg\min_{\phi_1', \phi_2'} \; \mathbb{E}_{\mathbb{P}_{\mathsf{G}}} r_{\mathrm{InfoBN}}(\mathsf{G}, \phi_1', \phi_2', \theta, \pi) \big\{ \mathcal{L}_{\mathrm{Gen}}(\mathsf{G}, \phi_1') \notag \\
     & + \mathcal{L}_{\mathrm{Gen}}(\mathsf{G}, \phi_2') \big\}, \ \   \pi \in \arg\min_{\pi'} \; \mathbb{E}_{\mathbb{P}_{\mathsf{G}}} \mathcal{L}_{\mathrm{InfoBN}}(\mathsf{G}, \phi_1, \phi_2, \theta, \pi').
\end{align}

\textbf{Incorporating InfoMin with InfoBN.}
We lastly incorporate InfoMin with InfoBN as Info(Min\&BN) to explore whether the proper collaboration of both principles can outperform each  individual.
We construct the collaborated reward dependent on the weighted summation of the estimated MI between two contrastive views and that between views and their embeddings, formulated as $r_{\text{Info(Min\&BN)}}(\mathsf{G}, \phi_1, \phi_2, \theta, \pi) =$ \\ $\Big\{ \begin{smallmatrix} 1, & \text{if} \; \gamma \mathcal{L}_{\mathrm{CL}}(\mathsf{G}, \phi_1, \phi_2, \theta) + (1-\gamma) \mathcal{L}_{\mathrm{InfoBN}}(\mathsf{G}, \phi_1, \phi_2, \theta, \pi) > \text{threshold} \\ \delta \ll 1, & \text{otherwise} \end{smallmatrix}$ with $\gamma \in [0, 1]$ as the major hyper-parameter tuned in Section \ref{sec:experiment}.

\begin{table*}[!htb]
\begin{center}
\caption{Semi-supervised learning on small-scale benchmarks from TUDataset (the first four) and large-scale ones from OGB (the last two).
Shown in \darkred{red} are the best three accuracies (\%) for TUDataset and the best
for ogbg-ppa and F1-score (\%) for ogbg-code.
The SOTA results compared here are as published under the same experimental setting (- indicates that results were not available in corresponding publications).}
\label{tab:semi_supervised}
\resizebox{0.8\textwidth}{!}{\begin{tabular}{c c c c c c c c}
    \hline
    \hline
    Methods & COLLAB & RDT-B & RDT-M5K & GITHUB & & ogbg-ppa & ogbg-code \\
    \hline
    \hline
    No pre-train & 73.71$\pm$0.27 & 86.63$\pm$0.27 & 51.33$\pm$0.44 & 60.87$\pm$0.17 & & 56.01$\pm$1.05 & 17.85$\pm$0.60 \\
    Augmentations & 74.19$\pm$0.13 & 87.74$\pm$0.39 & 52.01$\pm$0.20 & 60.91$\pm$0.32 & & - & - \\
    \hline
    GAE & \darkred{75.09}$\pm$0.19 & 87.69$\pm$0.40 & \darkred{53.58}$\pm$0.13 & \darkred{63.89}$\pm$0.52 & & - & - \\
    Infomax & 73.76$\pm$0.29 & \darkred{88.66}$\pm$0.95 & \darkred{53.61}$\pm$0.31 & \darkred{65.21}$\pm$0.88 & & - & - \\
    ContextPred & 73.69$\pm$0.37 & 84.76$\pm$0.52 & 51.23$\pm$0.84 & 62.35$\pm$0.73 & & - & -  \\
    GraphCL & 74.23$\pm$0.21 & \darkred{89.11}$\pm$0.19 & 52.55$\pm$0.45 & \darkred{65.81}$\pm$0.79 & & 57.77$\pm$1.25 & 22.45$\pm$0.17 \\
    \hline
    LP-InfoMin & \darkred{74.66}$\pm$0.14 & \darkred{88.03}$\pm$0.46 & 53.00$\pm$0.26 & 62.71$\pm$0.54 & & \darkred{59.10}$\pm$0.88 & 23.50$\pm$0.22 \\
    LP-InfoBN & 74.61$\pm$0.28 & 87.64$\pm$0.33 & 53.05$\pm$0.14 & 62.64$\pm$0.37 & & 55.48$\pm$0.97 & 23.31$\pm$0.22 \\
    LP-Info(Min\&BN) & \darkred{74.84}$\pm$0.31 & 87.81$\pm$0.45 & \darkred{53.32}$\pm$0.23 & 63.11$\pm$0.33 & & 57.31$\pm$0.99 & \darkred{23.61}$\pm$0.27 \\
    \hline
    \hline
\end{tabular}}
\end{center}
\end{table*}
\section{Experiments} \label{sec:experiment}
We evaluate our proposed method, GraphCL with learned prior (LP) against state-of-the-art (SOTA) competitors under the settings of semi-supervised learning and transfer learning  \cite{you2020graph,hu2019strategies}.
The use of other base models such as BGRL \cite{thakoor2021bootstrapped}  is included in Appendix \ref{appendx:base_model}.
The datasets include social networks, protein-protein interaction (PPI) networks, code abstract syntax trees and molecules, whose detailed statistics are available in \cite{Morris+2020,hu2020open,hu2019strategies}.
These data are released under the MIT license, and to our best knowledge, contain no privacy-infringing contents. 
We use the default hyper-parameters as in VGAE \cite{kipf2016variational} for the graph generative model except that for the encoder we adopt the same architecture as the pre-trained network due to the drastically diverse data formats across datasets. Without further optimization, we set $\delta=0.01$ and the threshold determined by the average condition value in the reward function (the estimated mutual information) in the corresponding batch and only tune $\gamma \in \{0.1, 0.3, 0.5, 0.7, 0.9\}$ in the Info(Min\&BN) principle based on validation.
We perform ablation studies in Appendix \ref{appendx:ablation}, and emphasize that by further tuning $\delta$ and the condition threshold, even stronger performances may be achieved by our models than reported in Section \ref{sec:mainresults}, as glimpsed through Section \ref{sec:hp_sensitivity}.  

Experiments are run on computer clusters with Tesla K80 GPU (11 GB memory) and NVIDIA A100 GPU (40 GB memory) and on personal servers with TITAN RTX (24 GB memory), GeForce RTX 2080 Ti (11 GB memory) and GeForce GTX 1080 Ti (11 GB memory). Numerical  efficiency is in Appendix \ref{appendx:efficiency}.  
Before detailing the results, we summarize main findings as follows: 
\vspace{-0.5em}
\begin{itemize}[leftmargin=*]
  \item In general, GraphCL with learned prior performs on par with the SOTA competitors on small benchmarks, and generalizes better on large-scale datasets, despite that it does not assume human expertise (reflected in  prefabricated augmentations or designed pretext tasks) and tedious trial-and-error based on downstream validation (Section \ref{sec:semi_supervised_learning} and \ref{sec:transfer_learning}).
  \item On the datasets with rigid/rich domain knowledge,  
  GraphCL with learned prior still outperforms that with the prefabricated one. Although it sometimes underperforms the predictive pretext tasks designed by human expertise, due to the overly-challenging knowledge learning purely from the data, it shows robust performance across datasets (Section \ref{sec:transfer_learning} and \ref{sec:vgae_vs_graphaf}).
  \item With more sufficient principled training to better capture knowledge from data, the learned prior with better quality usually leads to stronger performance (Section \ref{sec:semi_supervised_learning} and \ref{sec:correlation}).
\end{itemize}
\vspace{-0.5em}

\subsection{Comparison with the State of the Art}
\label{sec:mainresults}
\subsubsection{Semi-Supervised Learning} \label{sec:semi_supervised_learning}
\textbf{Setup.}
Four small social-network benchmarks are collected from TUDataset \cite{Morris+2020} and two large-scale graph datasets, protein-protein interaction networks and code abstract syntax trees, are gathered from Open Graph Benchmark (OGB) \cite{hu2020open}.
(i) For TUDataset benchmarks without explicit training/validation/test split, we perform self-supervised pre-training with all data and then  supervised fine-tuning/evaluation with 10 folds, each of which contains 10\% of data.
This whole procedure is repeated for 5 times to report the average performance and the error bar.
Residual graph convolutional network (ResGCN) \cite{chen2019powerful} is used as the backbone architecture with 5 layers and 128 hidden dimensions following GraphCL.
(ii) For OGB datasets with training/validation/test split, we perform self-supervised pre-training with all training data and supervised fine-tuning with 10\% of them then evaluate on the validation/test sets, which is repeated for 10 times.
Graph isomorphism network (GIN) \cite{xu2018powerful} is used with 5 layers and 300 hidden dimensions following \cite{hu2020open}.

\textbf{Results.}
 Table \ref{tab:semi_supervised} leads to the following observations.

\textbf{(i) On small benchmarks, learned prior performs on par with GraphCL and other SOTA methods.}
With the parameterized prior that is adaptively, dynamically and in principle learned from data, GraphCL with learned priors achieves better performance on COLLAB ($\ge$+0.38\%) and RDT-M5K ($\ge$+0.45\%) compared to GraphCL with hand-crafted augmentations. It has slightly worse results than GraphCL on RDT-B and GITHUB but the best-performing version is still comparable with other SOTAs (ranked the 3rd and 4th out of 9, respectively).
Similar observations are made in the classical small benchmarks (Appendix \ref{appendx:small_bench}).
Note that GraphCL 
selects augmentations by exhaustively and manually tuning on TUDataset \cite{you2020graph},
whereas our models are completely free from such labor. Therefore the competitive performance of our methods demonstrates the effectiveness of the learnable prior.  Interestingly, GAE performs well for these 4 benchmarks, showing its strong capability of learning priors for small-scale and regularly-structured social networks.

\textbf{(ii) On large-scale datasets, learned prior with appropriate principle(s) generalizes better versus pre-defined augmentations.}
Under the guidance of the proper principle, the learned prior leads to  better generalizability on ogbg-ppa (+1.33\%) and ogbg-code (+1.16\%) against manually designed and tuned augmentations in GraphCL.
This indicates that the data-driven learnable prior defined in a continuous space, rather than the manual selection from a discrete pool of pre-defined augmentations, can effectively evolve into the generalizable and scalable prior that  benefits downstream performance, even for large-scale graphs.

\begin{table*}[!htb]
\begin{center}
\caption{Transfer learning on molecular datasets from \cite{hu2019strategies}.
\darkred{Red} numbers indicate the best performances (AUROC, \%) and those within the standard deviation of the best. We compare with the published results.}
\label{tab:transfer_molecule}
\resizebox{0.9\textwidth}{!}{
\begin{tabular}{c c c c c c c c c }
    \hline
    \hline
    Methods & BBBP & Tox21 & ToxCast & SIDER & ClinTox & MUV & HIV & BACE \\
    \hline
    \hline
    No pre-train. & 65.8$\pm$4.5 & 74.0$\pm$0.8 & 63.4$\pm$0.6 & 57.3$\pm$1.6 & 58.0$\pm$4.4 & 71.8$\pm$2.5 & 75.3$\pm$1.9 & 70.1$\pm$5.4 \\
    \hline
    Infomax & 68.8$\pm$0.8 & 75.3$\pm$0.5 & 62.7$\pm$0.4 & 58.4$\pm$0.8 & 69.9$\pm$3.0 & \darkred{75.3}$\pm$2.5 & 76.0$\pm$0.7 & 75.9$\pm$1.6 \\
    EdgePred & 67.3$\pm$2.4 & 76.0$\pm$0.6 & \darkred{64.1}$\pm$0.6 & \darkred{60.4}$\pm$0.7 & 64.1$\pm$3.7 & 74.1$\pm$2.1 & 76.3$\pm$1.0 & 79.9$\pm$0.9 \\
    AttrMasking & 64.3$\pm$2.8 & \darkred{76.7}$\pm$0.4 & \darkred{64.2}$\pm$0.5 & \darkred{61.0}$\pm$0.7 & 71.8$\pm$4.1 & \darkred{74.7}$\pm$1.4 & 77.2$\pm$1.1 & 79.3$\pm$1.6 \\
    ContextPred & 68.0$\pm$2.0 & 75.7$\pm$0.7 & \darkred{63.9}$\pm$0.6 & \darkred{60.9}$\pm$0.6 & 65.9$\pm$3.8 & \darkred{75.8}$\pm$1.7 & \darkred{77.3}$\pm$1.0 & 79.6$\pm$1.2 \\
    GraphCL & 69.68$\pm$0.67 & 73.87$\pm$0.66 & 62.40$\pm$0.57 & \darkred{60.53}$\pm$0.88 & \darkred{75.99}$\pm$2.65 & 69.80$\pm$2.66 & \darkred{78.47}$\pm$1.22 & 75.38$\pm$1.44 \\
    \hline
    LP-InfoMin & \darkred{71.47}$\pm$0.66 & 74.60$\pm$0.70 & 63.13$\pm$0.30 & \darkred{60.52}$\pm$0.75 & 72.39$\pm$1.50 & 70.51$\pm$2.25 & 76.43$\pm$0.85 & 78.86$\pm$1.66 \\
    LP-InfoBN & \darkred{71.68}$\pm$0.99 & 74.45$\pm$0.51 & 62.39$\pm$0.60 & \darkred{60.80}$\pm$0.62 & \darkred{76.73}$\pm$1.53 & 72.03$\pm$1.17 & 77.03$\pm$0.75 & \darkred{81.15}$\pm$1.33 \\
    LP-Info(Min\&BN) & \darkred{71.40}$\pm$0.55 & 74.54$\pm$0.45 & 63.04$\pm$0.30 & 59.70$\pm$0.43 & 74.81$\pm$2.73 & 72.99$\pm$2.28 & 76.96$\pm$1.10 & \darkred{80.21}$\pm$1.36 \\
    \hline
    \hline
\end{tabular}}
\end{center}
\end{table*}
\textbf{(iii) Rule of choosing principle is dataset-dependent, and reconstruction quality is a potential indicator.}
Across different datasets, we observe that the benefit of various principles (InfoMin, InfoBN, or both) to guide prior learning is dataset-dependent, which is intuitively reasonable given the enormously diverse nature of graph datasets. 
Toward a rule of choosing the proper principle again without relying on  downstream validation, we find that the reconstruction quality of the graph generator can potentially be an indicator, which will be discussed in Section \ref{sec:correlation}.


\subsubsection{Transfer Learning} \label{sec:transfer_learning}
\textbf{Setup.}
One pre-training dataset for molecules and eight fine-tuning/evaluation datasets are taken from ZINC15 \cite{sterling2015zinc} and MoleculeNet \cite{wu2018moleculenet,hu2019strategies}, respectively.  
And one pre-training dataset for protein-protein interaction networks and one fine-tuning/ evaluation dataset are curated in \cite{hu2019strategies}.
We pre-train on the larger dataset then fine-tune/evaluate on smaller datasets of the same category using the given training/validation/test split.
GIN is used with 5 layers and 300 hidden dimensions following \cite{hu2019strategies}.

\textbf{Results.}
 Table \ref{tab:transfer_molecule} and \ref{tab:transfer_ppi}, reveal the following observations.

\textbf{(iv) On datasets with rich/rigid knowledge, learned prior still beats contrastive learning with pre-fabricated augmentations but can underperform self-supervision with expert-designed pretext tasks.}
On the molecular datasets, the learned prior with proper principle(s) achieves significant improvement compared to GraphCL, boosting 7 of 8 downstream performances (Table \ref{tab:transfer_molecule}).
This again echos our previous observation on relational graphs that the learnable prior results in  better generalizability and scalability.
Meanwhile, it can underperform the SOTA self-supervised methods with pretext tasks carefully designed for the molecule domain. Nevertheless, the best-performing learnable prior still manages to outperform other SOTA methods for 3 out of 8 datasets.  
\begin{table}[!htb]
\begin{center}
 \caption{Transfer learning on PPI dataset from \cite{hu2019strategies}.
\darkred{Red} numbers indicate the best performances (AUROC, \%) and those within the standard deviation of the best.
The compared results are from the published papers.}
\label{tab:transfer_ppi}
\resizebox{0.25\textwidth}{!}{\begin{tabular}{c c c c c c c c c c }
    \hline
    \hline
    Methods & PPI \\
    \hline
    \hline
    No pre-train. & 64.8$\pm$1.0 \\
    \hline
    Infomax & 64.1$\pm$1.5 \\
    EdgePred & 65.7$\pm$1.3 \\
    AttrMasking & 65.2$\pm$1.6 \\
    ContextPred & 64.4$\pm$1.3 \\
    GraphCL & 67.88$\pm$0.85 \\
    \hline
    LP-InfoMin & 68.24$\pm$0.87 \\
    LP-InfoBN & \darkred{71.16}$\pm$0.28 \\
    LP-Info(Min\&BN) & 70.10$\pm$0.76 \\
    \hline
    \hline
\end{tabular}}
\end{center}
\end{table}

Unlike relational networks (e.g. citation networks earlier and PPI next) \cite{kipf2016semi}, molecules are required to follow rich and ``rigid'' chemistry rules to be semantically valid and their casual ``augmentations''  (e.g. modifying nodes/atoms or edges/bonds) can lead to invalidity,
which is much more challenging for the generator to explore directly from data.
Indeed, examples in \cite{hu2019strategies,you2020graph} show that improper domain-naive pretext tasks or augmentations lead to performance degradation for molecules; whereas some pretext tasks carefully designed to meet domain knowledge and human expertise boost downstream performance.  Nevertheless, without assumption of the knowledge and the downstream labels, our learned prior has demonstrated robust competitiveness across datasets.   

On the other hand, for the PPI dataset (and those in Section \ref{sec:semi_supervised_learning}) where graphs are less embedded with the  richness and/or rigidness of domain knowledge, we show the clear advantage of the learned prior that achieves the best performance ($\ge$+3.28\%) compared with all other SOTA competitors in Table \ref{tab:transfer_ppi}.
This again demonstrates the effectiveness of learned prior.

\begin{table*}[!htb]
\begin{center}
\caption{Link prediction performance (AUROC and AUPRC, \%) of VGAE generators on eight pre-training datasets. Better link prediction results are marked in \darkred{red} if accompanied with better downstream performances, as shown in Table \ref{tab:semi_supervised}, \ref{tab:transfer_molecule} and \ref{tab:transfer_ppi}.}
\label{tab:correlation}
\resizebox{0.85\textwidth}{!}{\begin{tabular}{c c c c c c c c c c}
    \hline
    \hline
    & Principles & COLLAB & RDT-B & RDT-M5K & GITHUB & ogbg-ppa & ogbg-code & Trans-Mol & Trans-PPI \\
    \hline
    \hline
    \multirow{2}{*}{AUROC (\%)} & InfoMin & \darkred{71.28} & \darkred{97.32} & 99.08 & 78.68 & \darkred{96.53} & 92.39 & 64.54 & 71.20 \\
    & InfoBN & 69.44 & 97.29 & \darkred{99.31} & 81.20 & 95.24 & 94.06 & \darkred{83.55} & \darkred{71.32} \\
    \hline
    \multirow{2}{*}{AUPRC (\%)} & InfoMin & \darkred{80.84} & \darkred{96.62} & 98.67 & 78.49 & \darkred{95.94} & 90.08 & 64.14 & 69.34 \\
    & InfoBN & 79.13 & 96.59 & \darkred{98.97} & 80.47 & 95.30 & 91.66 & \darkred{82.51} & \darkred{70.65} \\
    \hline
    \hline
\end{tabular}}
\end{center}
\end{table*}

\begin{figure*}[!htb]
  \begin{center}
    \includegraphics[width=0.85\textwidth]{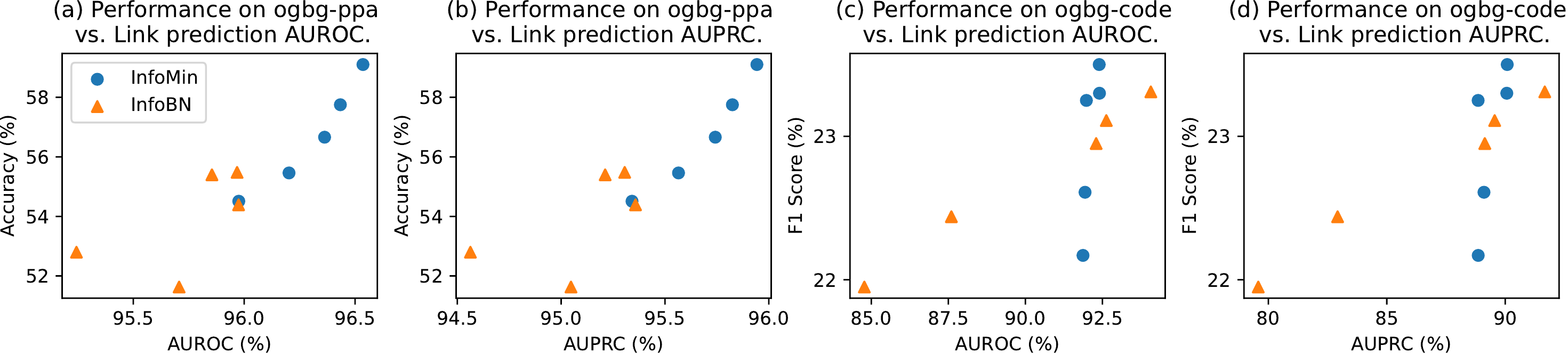}
  \end{center}
  \caption{Link prediction performance (AUROC and AUPRC, \%) of the generator checkpoints on two representative large-scale datastes ogbg-ppa and ogbg-code.}
  \label{fig:correlation}
\end{figure*}

\subsection{Further Analyses}
With performance comparison with SOTA methods completed, we  delve into the proposed models for even more insights. 
\subsubsection{Graph generation quality usually aligns with downstream performance.} \label{sec:correlation}
The first question we would like to answer is:  
what is the relationship between the quality of the learned graph generative model (specifically VGAE here) and the downstream performance?
In the past research, it is widely adopted to evaluate VGAE using its reconstruction quality, which can be quantitatively measured by the performance of link prediction, a surrogate task  \cite{kipf2016variational,zhang2018link,wang2018graphgan}.
We employ this commonly-used surrogate evaluation and find out that in most cases better reconstruction quality aligns with better downstream performance, as shown below.

Table \ref{tab:correlation}  shows that under the guidance of the InfoMin and InfoBN principles, the generator makes more precise link  prediction (in both AUROC and AUPRC), leading to better downstream performance for 6 out of the 8  datasets.
The same trend is observed among different checkpoints during training, as seen in Figure \ref{fig:correlation}.

This observation further sheds light on the following insights.
(i) The better reconstruction quality of the generator stands for more sufficient principled training to capture the prior knowledge, leading to  better performance.
Thus, the quality
of the learned prior is inherently related to the  generalizability of graph contrastive learning.
This acts as the causal factor of the performance in most cases.
(ii) Furthermore, the alignment offers a potential criterion for selecting the principle and early stopping while learning the prior (as stated in Section \ref{sec:semi_supervised_learning}), without the need of human intervention or downstream validation.

\subsubsection{Molecule-specific generator alone does not significantly  benefit molecular datasets.} \label{sec:vgae_vs_graphaf}
Next we explore whether a domain-compatible molecule-specific generator, rather than the domain-agnostic graph generative model VGAE, can further boost the performance in Section \ref{sec:transfer_learning} and potentially meet or beat expert-designed pretext tasks in all 8 datasets.
As mentioned earlier, it might be too challenging for VGAE without  domain-adaptation to learn the rigid/rich domain knowledge of molecules purely from data.  
Therefore, we replace VGAE with GraphAF \cite{shi2020graphaf} which is not only molecule-specific but also sample-efficient compared with other molecule generators \cite{jin2018junction,liu2021graphebm,you2018graphrnn}.
Efficiency is crucial in our case since sampling is performed in each feed-forward propagation of the framework \eqref{eq:graphcl_learned_prior_reward}.
We show the results in Table \ref{tab:vgae_vs_graphaf} with the InfoMin principle incorporated.
\begin{table*}[!htb]
\begin{center}
\caption{Learned prior performance with different generators under the guidance of InfoMin, in the transfer learning setting on molecular datasets.
\darkred{Red} numbers indicate the best performances (AUROC, \%).}
\label{tab:vgae_vs_graphaf}
\resizebox{0.85\textwidth}{!}{
\begin{tabular}{c c c c c c c c c }
    \hline
    \hline
    Methods & BBBP & Tox21 & ToxCast & SIDER & ClinTox & MUV & HIV & BACE \\
    \hline
    VGAE & \darkred{71.47}$\pm$0.66 & \darkred{74.60}$\pm$0.70 & \darkred{63.13}$\pm$0.30 & 60.52$\pm$0.75 & 72.39$\pm$1.50 & 70.51$\pm$2.25 & \darkred{76.43}$\pm$0.85 & \darkred{78.86}$\pm$1.66 \\
    GraphAF & 70.55$\pm$0.63 & 73.51$\pm$0.43 & 62.03$\pm$0.33 & \darkred{61.32}$\pm$1.32 & \darkred{77.47}$\pm$1.91 & \darkred{72.25}$\pm$1.18 & 76.30$\pm$1.34 & 78.43$\pm$2.36 \\
    \hline
    \hline
\end{tabular}}
\end{center}
\end{table*}

The results above do not show a significant performance difference between the two graph generators.  We believe that additional work 
is needed to bring out the potential benefit of domain-specific knowledge-infused graph generators.  

\subsubsection{Tuning principled-reward hyper-parameters could  strengthen the competitive performance even more.} \label{sec:hp_sensitivity}
\begin{figure*}[!htb]
  \begin{center}
    \includegraphics[width=0.85\textwidth]{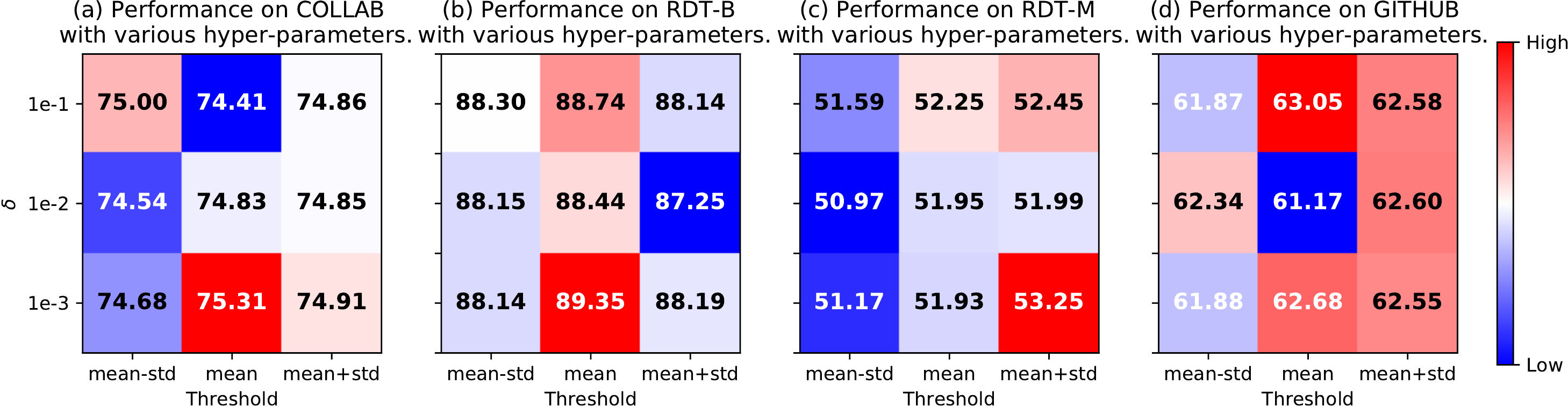}
  \end{center}
  \caption{Impact of hyper-parameters of the reward function for InfoMin,  $\delta$ and threshold, on COLLAB, RDT-B, RDT-M and GITHUB datasets. Warmer colors represent the relatively better performances.} \label{fig:sensitivity}
\end{figure*}

Furthermore, we would like to examine the influence of two unique hyper-parameters in the reward function of the framework \eqref{eq:graphcl_learned_prior_reward}, $\delta$ values and the threshold determination, which we did not have the luxury to tune in earlier results. Previously we fix $\delta=0.01$ and set the threshold at the mean value (of the corresponding estimated mutual information in the batch) due to the limited computational resources and the large amount of experiments.
Here on four small-scale benchmarks, COLLAB, RDT-B, RDT-M and GITHUB from the TUDateset, with the InfoMin principle, we tune $(\delta, \; \text{threshold}) \in \{0.1, 0.01, 0.001\} \times \{\text{mean-sd, mean, mean+sd}\}$ (sd: standard deviation) while fixing the optimal $\gamma$, with results shown in Figure \ref{fig:sensitivity}.

Results echo the similar observation as in Section \ref{sec:semi_supervised_learning} that, the choice of the hyper-parameter $(\delta, \; \text{threshold})$ is dataset-relevant owing to the distinctively heterogeneous nature of graph data.
We notice that earlier chosen values $(\delta, \; \text{threshold}) = (0.01, \; \text{mean})$ located in the middle patch of heatmaps in Figure \ref{fig:sensitivity}, are not optimal (even not within top-3 performance on 3 out of 4 datasets), whereas our models' performances further improve their competitiveness against SOTA methods.  

\begin{figure*}[!htb]
  \begin{center}
    \includegraphics[width=0.75\textwidth]{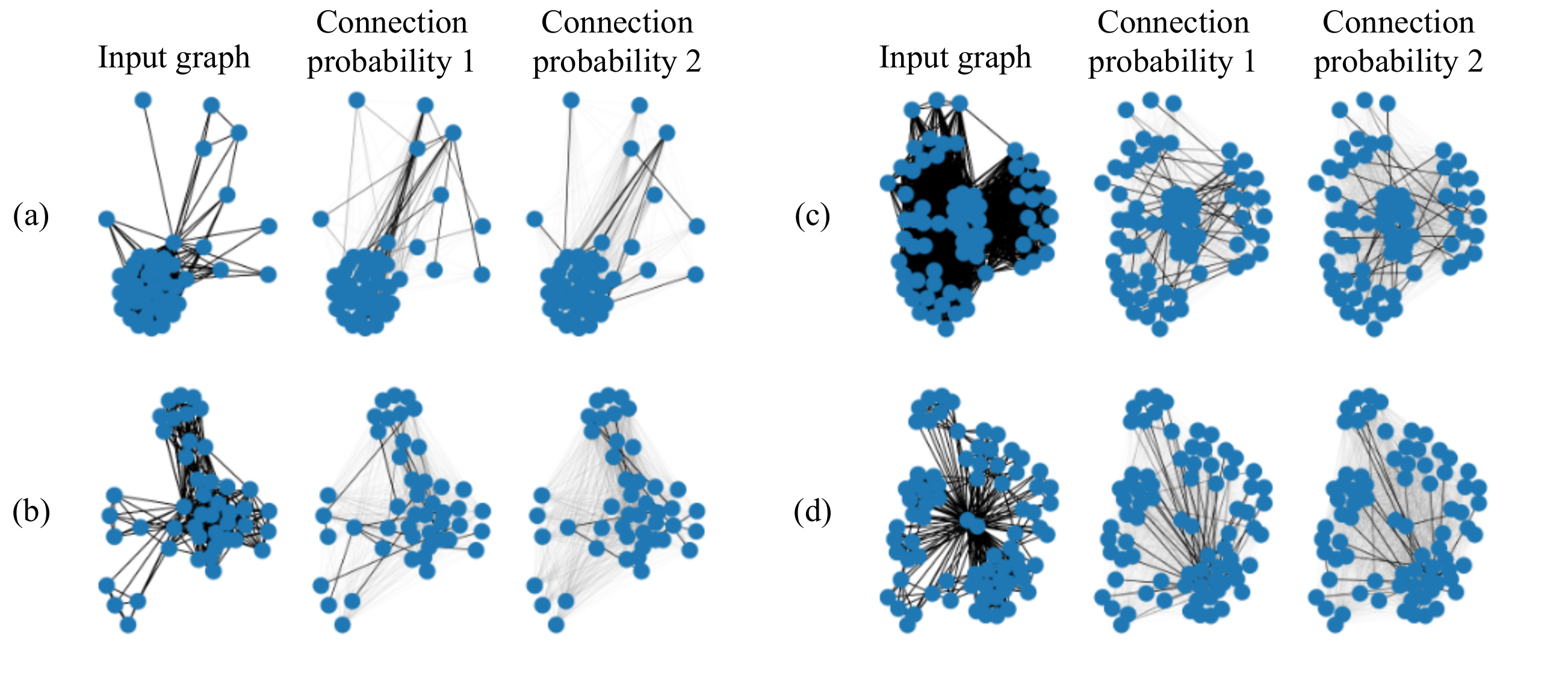}
  \end{center} 
  \caption{Visualizing the connection probability of generative models guided by Info(Min\&BN) for 4 samples from COLLAB.} \label{fig:viz_graph}
\end{figure*} 

\subsubsection{Generative graphs connections are sparse to capture patterns.}
We lastly visualize the generated connections probability under the guidance of Info(Min\&BN) in Figure \ref{fig:viz_graph}.
Compared with input graphs, we observe the sparsity in the generated connections, that tries to preserve links among cliques and some key connection across cliques.
This sparsification is useful to capture patterns of original graphs and generate contrastive views.
We leave the quantitive assessment of the ``usefulness" in future work.


\section{Conclusions}\label{sec:conclusion}
In this paper, we target more adaptive, automatic and generalizable graph self-supervised learning, by introducing a learnable prior and a framework to learn it.  
Leveraging the SOTA GraphCL framework as the base model, we extend the prefabricated discrete prior into a learnable continuous one parameterized by a graph generative model.
In addition, principles such as InfoMin and InfoBN are incorporated to avoid collapsing into trivial solutions.
The resulting framework is formulated as  bi-level optimization.
Empirically, this first attempt to incorporate the adaptive and dynamic learned prior with GNN, performs on par with the SOTA competitors on small benchmarks, and generalizes better on large-scale datasets, without resorting to human expertise of domain knowledge or tedious trial-and-error relying on downstream validation.

Our proposed learnable prior further exploits the power of deep learning and big data in the graph domain, and therefore is of broad interests and wide applications as in recommendation systems, drug discovery and combinatorial optimization.

\section*{Acknowledgment}
The study is in part funded by NIH (R35GM124952 to YS). 




\bibliographystyle{unsrt}
\bibliography{acmart}

\begin{thebibliography}{10}

\bibitem{xie2021self}
Yaochen Xie, Zhao Xu, Zhengyang Wang, and Shuiwang Ji.
\newblock Self-supervised learning of graph neural networks: A unified review.
\newblock {\em arXiv preprint arXiv:2102.10757}, 2021.

\bibitem{liu2021graph}
Yixin Liu, Shirui Pan, Ming Jin, Chuan Zhou, Feng Xia, and Philip~S Yu.
\newblock Graph self-supervised learning: A survey.
\newblock {\em arXiv preprint arXiv:2103.00111}, 2021.

\bibitem{hu2019strategies}
Weihua Hu, Bowen Liu, Joseph Gomes, Marinka Zitnik, Percy Liang, Vijay Pande,
  and Jure Leskovec.
\newblock Strategies for pre-training graph neural networks.
\newblock {\em arXiv preprint arXiv:1905.12265}, 2019.

\bibitem{you2020graph}
Yuning You, Tianlong Chen, Yongduo Sui, Ting Chen, Zhangyang Wang, and Yang
  Shen.
\newblock Graph contrastive learning with augmentations.
\newblock {\em Advances in Neural Information Processing Systems}, 33, 2020.

\bibitem{liu2020self}
Xiao Liu, Fanjin Zhang, Zhenyu Hou, Zhaoyu Wang, Li~Mian, Jing Zhang, and Jie
  Tang.
\newblock Self-supervised learning: Generative or contrastive.
\newblock {\em arXiv preprint arXiv:2006.08218}, 1(2), 2020.

\bibitem{kipf2016semi}
Thomas~N Kipf and Max Welling.
\newblock Semi-supervised classification with graph convolutional networks.
\newblock {\em arXiv preprint arXiv:1609.02907}, 2016.

\bibitem{zou2019layer}
Difan Zou, Ziniu Hu, Yewen Wang, Song Jiang, Yizhou Sun, and Quanquan Gu.
\newblock Layer-dependent importance sampling for training deep and large graph
  convolutional networks.
\newblock {\em arXiv preprint arXiv:1911.07323}, 2019.

\bibitem{you2020cross}
Yuning You and Yang Shen.
\newblock Cross-modality protein embedding for compound-protein affinity and
  contact prediction.
\newblock {\em arXiv preprint arXiv:2012.00651}, 2020.

\bibitem{you2020l2}
Yuning You, Tianlong Chen, Zhangyang Wang, and Yang Shen.
\newblock L2-gcn: Layer-wise and learned efficient training of graph
  convolutional networks.
\newblock In {\em Proceedings of the IEEE/CVF Conference on Computer Vision and
  Pattern Recognition}, pages 2127--2135, 2020.

\bibitem{hwang2020self}
Dasol Hwang, Jinyoung Park, Sunyoung Kwon, Kyung-Min Kim, Jung-Woo Ha, and
  Hyunwoo~J Kim.
\newblock Self-supervised auxiliary learning with meta-paths for heterogeneous
  graphs.
\newblock {\em arXiv preprint arXiv:2007.08294}, 2020.

\bibitem{you2020does}
Yuning You, Tianlong Chen, Zhangyang Wang, and Yang Shen.
\newblock When does self-supervision help graph convolutional networks?
\newblock In {\em International Conference on Machine Learning}, pages
  10871--10880. PMLR, 2020.

\bibitem{perozzi2014deepwalk}
Bryan Perozzi, Rami Al-Rfou, and Steven Skiena.
\newblock Deepwalk: Online learning of social representations.
\newblock In {\em Proceedings of the 20th ACM SIGKDD international conference
  on Knowledge discovery and data mining}, pages 701--710, 2014.

\bibitem{tang2015line}
Jian Tang, Meng Qu, Mingzhe Wang, Ming Zhang, Jun Yan, and Qiaozhu Mei.
\newblock Line: Large-scale information network embedding.
\newblock In {\em Proceedings of the 24th international conference on world
  wide web}, pages 1067--1077, 2015.

\bibitem{kipf2016variational}
Thomas~N Kipf and Max Welling.
\newblock Variational graph auto-encoders.
\newblock {\em arXiv preprint arXiv:1611.07308}, 2016.

\bibitem{hamilton2017inductive}
William~L Hamilton, Rex Ying, and Jure Leskovec.
\newblock Inductive representation learning on large graphs.
\newblock {\em arXiv preprint arXiv:1706.02216}, 2017.

\bibitem{hu2020gpt}
Ziniu Hu, Yuxiao Dong, Kuansan Wang, Kai-Wei Chang, and Yizhou Sun.
\newblock {GPT-GNN}: Generative pre-training of graph neural networks.
\newblock In {\em Proceedings of the 26th ACM SIGKDD International Conference
  on Knowledge Discovery \& Data Mining}, pages 1857--1867, 2020.

\bibitem{sun2020multi}
Ke~Sun, Zhouchen Lin, and Zhanxing Zhu.
\newblock Multi-stage self-supervised learning for graph convolutional networks
  on graphs with few labeled nodes.
\newblock In {\em Proceedings of the AAAI Conference on Artificial
  Intelligence}, volume~34, pages 5892--5899, 2020.

\bibitem{sehanobish2020self}
Arijit Sehanobish, Neal~G Ravindra, and David van Dijk.
\newblock Self-supervised edge features for improved graph neural network
  training.
\newblock {\em arXiv preprint arXiv:2007.04777}, 2020.

\bibitem{jin2020self}
Wei Jin, Tyler Derr, Haochen Liu, Yiqi Wang, Suhang Wang, Zitao Liu, and
  Jiliang Tang.
\newblock Self-supervised learning on graphs: Deep insights and new direction.
\newblock {\em arXiv preprint arXiv:2006.10141}, 2020.

\bibitem{rong2020self}
Yu~Rong, Yatao Bian, Tingyang Xu, Weiyang Xie, Ying Wei, Wenbing Huang, and
  Junzhou Huang.
\newblock Self-supervised graph transformer on large-scale molecular data.
\newblock {\em Advances in Neural Information Processing Systems}, 33, 2020.

\bibitem{hao2021pre}
Bowen Hao, Jing Zhang, Hongzhi Yin, Cuiping Li, and Hong Chen.
\newblock Pre-training graph neural networks for cold-start users and items
  representation.
\newblock In {\em Proceedings of the 14th ACM International Conference on Web
  Search and Data Mining}, pages 265--273, 2021.

\bibitem{kim2021find}
Dongkwan Kim and Alice Oh.
\newblock How to find your friendly neighborhood: Graph attention design with
  self-supervision.
\newblock In {\em International Conference on Learning Representations}, 2021.

\bibitem{yu2021self}
Junliang Yu, Hongzhi Yin, Jundong Li, Qinyong Wang, Nguyen Quoc~Viet Hung, and
  Xiangliang Zhang.
\newblock Self-supervised multi-channel hypergraph convolutional network for
  social recommendation.
\newblock {\em arXiv preprint arXiv:2101.06448}, 2021.

\bibitem{zhang2020iterative}
Hanlin Zhang, Shuai Lin, Weiyang Liu, Pan Zhou, Jian Tang, Xiaodan Liang, and
  Eric~P Xing.
\newblock Iterative graph self-distillation.
\newblock {\em arXiv preprint arXiv:2010.12609}, 2020.

\bibitem{hwang2021self}
Dasol Hwang, Jinyoung Park, Sunyoung Kwon, Kyung-Min Kim, Jung-Woo Ha, et~al.
\newblock Self-supervised auxiliary learning for graph neural networks via
  meta-learning.
\newblock {\em arXiv preprint arXiv:2103.00771}, 2021.

\bibitem{li2021representation}
Michelle~M Li, Kexin Huang, and Marinka Zitnik.
\newblock Representation learning for networks in biology and medicine:
  Advancements, challenges, and opportunities.
\newblock {\em arXiv preprint arXiv:2104.04883}, 2021.

\bibitem{huang2021hop}
Tianjin Huang, Yulong Pei, Vlado Menkovski, and Mykola Pechenizkiy.
\newblock Hop-count based self-supervised anomaly detection on attributed
  networks.
\newblock {\em arXiv preprint arXiv:2104.07917}, 2021.

\bibitem{manessi2021graph}
Franco Manessi and Alessandro Rozza.
\newblock Graph-based neural network models with multiple self-supervised
  auxiliary tasks.
\newblock {\em Pattern Recognition Letters}, 2021.

\bibitem{suresh2021adversarial}
Susheel Suresh, Pan Li, Cong Hao, and Jennifer Neville.
\newblock Adversarial graph augmentation to improve graph contrastive learning.
\newblock {\em arXiv preprint arXiv:2106.05819}, 2021.

\bibitem{xu2021infogcl}
Dongkuan Xu, Wei Cheng, Dongsheng Luo, Haifeng Chen, and Xiang Zhang.
\newblock Infogcl: Information-aware graph contrastive learning.
\newblock {\em Advances in Neural Information Processing Systems}, 34, 2021.

\bibitem{kefatoself}
Zekarias~T Kefato, Sarunas Girdzijauskas, and Hannes St{\"a}rk.
\newblock Self-supervised gnn that jointly learns to augment.

\bibitem{xu2021group}
Xinyi Xu, Cheng Deng, Yaochen Xie, and Shuiwang Ji.
\newblock Group contrastive self-supervised learning on graphs.
\newblock {\em arXiv preprint arXiv:2107.09787}, 2021.

\bibitem{sun2019infograph}
Fan-Yun Sun, Jordan Hoffmann, Vikas Verma, and Jian Tang.
\newblock Infograph: Unsupervised and semi-supervised graph-level
  representation learning via mutual information maximization.
\newblock {\em arXiv preprint arXiv:1908.01000}, 2019.

\bibitem{hassani2020contrastive}
Kaveh Hassani and Amir~Hosein Khasahmadi.
\newblock Contrastive multi-view representation learning on graphs.
\newblock {\em arXiv preprint arXiv:2006.05582}, 2020.

\bibitem{qiu2020gcc}
Jiezhong Qiu, Qibin Chen, Yuxiao Dong, Jing Zhang, Hongxia Yang, Ming Ding,
  Kuansan Wang, and Jie Tang.
\newblock Gcc: Graph contrastive coding for graph neural network pre-training.
\newblock In {\em Proceedings of the 26th ACM SIGKDD International Conference
  on Knowledge Discovery \& Data Mining}, 2020.

\bibitem{velivckovic2018deep}
Petar Veli{\v{c}}kovi{\'c}, William Fedus, William~L Hamilton, Pietro Li{\`o},
  Yoshua Bengio, and R~Devon Hjelm.
\newblock Deep graph infomax.
\newblock {\em arXiv preprint arXiv:1809.10341}, 2018.

\bibitem{peng2020self}
Zhen Peng, Yixiang Dong, Minnan Luo, Xiao-Ming Wu, and Qinghua Zheng.
\newblock Self-supervised graph representation learning via global context
  prediction.
\newblock {\em arXiv preprint arXiv:2003.01604}, 2020.

\bibitem{zhu2020graph}
Yanqiao Zhu, Yichen Xu, Feng Yu, Qiang Liu, Shu Wu, and Liang Wang.
\newblock Graph contrastive learning with adaptive augmentation.
\newblock {\em arXiv preprint arXiv:2010.14945}, 2020.

\bibitem{chen2020distance}
Deli Chen, Yanyai Lin, Lei Li, Xuancheng~Ren Li, Jie Zhou, Xu~Sun, et~al.
\newblock Distance-wise graph contrastive learning.
\newblock {\em arXiv preprint arXiv:2012.07437}, 2020.

\bibitem{chen2020coad}
Bo~Chen, Jing Zhang, Xiaokang Zhang, Xiaobin Tang, Lingfan Cai, Hong Chen,
  Cuiping Li, Peng Zhang, and Jie Tang.
\newblock Coad: Contrastive pre-training with adversarial fine-tuning for
  zero-shot expert linking.
\newblock {\em arXiv preprint arXiv:2012.11336}, 2020.

\bibitem{ren2019heterogeneous}
Yuxiang Ren, Bo~Liu, Chao Huang, Peng Dai, Liefeng Bo, and Jiawei Zhang.
\newblock Heterogeneous deep graph infomax.
\newblock {\em arXiv preprint arXiv:1911.08538}, 2019.

\bibitem{park2020unsupervised}
Chanyoung Park, Donghyun Kim, Jiawei Han, and Hwanjo Yu.
\newblock Unsupervised attributed multiplex network embedding.
\newblock In {\em AAAI}, pages 5371--5378, 2020.

\bibitem{peng2020graph}
Zhen Peng, Wenbing Huang, Minnan Luo, Qinghua Zheng, Yu~Rong, Tingyang Xu, and
  Junzhou Huang.
\newblock Graph representation learning via graphical mutual information
  maximization.
\newblock In {\em Proceedings of The Web Conference 2020}, pages 259--270,
  2020.

\bibitem{zhang2020motif}
Shichang Zhang, Ziniu Hu, Arjun Subramonian, and Yizhou Sun.
\newblock Motif-driven contrastive learning of graph representations.
\newblock {\em arXiv preprint arXiv:2012.12533}, 2020.

\bibitem{roy2021node}
Kashob~Kumar Roy, Amit Roy, AKM Rahman, M~Ashraful Amin, and Amin~Ahsan Ali.
\newblock Node embedding using mutual information and self-supervision based
  bi-level aggregation.
\newblock {\em arXiv preprint arXiv:2104.13014}, 2021.

\bibitem{zhao2020data}
Tong Zhao, Yozen Liu, Leonardo Neves, Oliver Woodford, Meng Jiang, and Neil
  Shah.
\newblock Data augmentation for graph neural networks.
\newblock {\em arXiv preprint arXiv:2006.06830}, 2020.

\bibitem{kong2020flag}
Kezhi Kong, Guohao Li, Mucong Ding, Zuxuan Wu, Chen Zhu, Bernard Ghanem, Gavin
  Taylor, and Tom Goldstein.
\newblock Flag: Adversarial data augmentation for graph neural networks.
\newblock {\em arXiv preprint arXiv:2010.09891}, 2020.

\bibitem{verma2019graphmix}
Vikas Verma, Meng Qu, Alex Lamb, Yoshua Bengio, Juho Kannala, and Jian Tang.
\newblock Graphmix: Regularized training of graph neural networks for
  semi-supervised learning.
\newblock {\em arXiv preprint arXiv:1909.11715}, 2019.

\bibitem{jin2021automated}
Wei Jin, Xiaorui Liu, Xiangyu Zhao, Yao Ma, Neil Shah, and Jiliang Tang.
\newblock Automated self-supervised learning for graphs.
\newblock {\em arXiv preprint arXiv:2106.05470}, 2021.

\bibitem{you2021graph}
Yuning You, Tianlong Chen, Yang Shen, and Zhangyang Wang.
\newblock Graph contrastive learning automated.
\newblock {\em arXiv preprint arXiv:2106.07594}, 2021.

\bibitem{denton2018stochastic}
Emily Denton and Rob Fergus.
\newblock Stochastic video generation with a learned prior.
\newblock In {\em International Conference on Machine Learning}, pages
  1174--1183. PMLR, 2018.

\bibitem{bora2017compressed}
Ashish Bora, Ajil Jalal, Eric Price, and Alexandros~G Dimakis.
\newblock Compressed sensing using generative models.
\newblock In {\em International Conference on Machine Learning}, pages
  537--546. PMLR, 2017.

\bibitem{tripathi2018correction}
Subarna Tripathi, Zachary~C Lipton, and Truong~Q Nguyen.
\newblock Correction by projection: Denoising images with generative
  adversarial networks.
\newblock {\em arXiv preprint arXiv:1803.04477}, 2018.

\bibitem{grover2018amortized}
Aditya Grover and Stefano Ermon.
\newblock Amortized variational compressive sensing.
\newblock 2018.

\bibitem{kabkab2018task}
Maya Kabkab, Pouya Samangouei, and Rama Chellappa.
\newblock Task-aware compressed sensing with generative adversarial networks.
\newblock In {\em Proceedings of the AAAI Conference on Artificial
  Intelligence}, volume~32, 2018.

\bibitem{shah2018solving}
Viraj Shah and Chinmay Hegde.
\newblock Solving linear inverse problems using gan priors: An algorithm with
  provable guarantees.
\newblock In {\em 2018 IEEE international conference on acoustics, speech and
  signal processing (ICASSP)}, pages 4609--4613. IEEE, 2018.

\bibitem{fletcher2018inference}
Alyson~K Fletcher, Sundeep Rangan, and Philip Schniter.
\newblock Inference in deep networks in high dimensions.
\newblock In {\em 2018 IEEE International Symposium on Information Theory
  (ISIT)}, pages 1884--1888. IEEE, 2018.

\bibitem{asim2018solving}
Muhammad Asim, Fahad Shamshad, and Ali Ahmed.
\newblock Solving bilinear inverse problems using deep generative priors.
\newblock {\em CoRR, abs/1802.04073}, 3(4):8, 2018.

\bibitem{hand2018global}
Paul Hand and Vladislav Voroninski.
\newblock Global guarantees for enforcing deep generative priors by empirical
  risk.
\newblock In {\em Conference On Learning Theory}, pages 970--978. PMLR, 2018.

\bibitem{fortuin2021priors}
Vincent Fortuin.
\newblock Priors in bayesian deep learning: A review, 2021.

\bibitem{ulyanov2018deep}
Dmitry Ulyanov, Andrea Vedaldi, and Victor Lempitsky.
\newblock Deep image prior.
\newblock In {\em Proceedings of the IEEE conference on computer vision and
  pattern recognition}, pages 9446--9454, 2018.

\bibitem{chakrabarti2006graph}
Deepayan Chakrabarti and Christos Faloutsos.
\newblock Graph mining: Laws, generators, and algorithms.
\newblock {\em ACM computing surveys (CSUR)}, 38(1):2--es, 2006.

\bibitem{wang2018graphgan}
Hongwei Wang, Jia Wang, Jialin Wang, Miao Zhao, Weinan Zhang, Fuzheng Zhang,
  Xing Xie, and Minyi Guo.
\newblock Graphgan: Graph representation learning with generative adversarial
  nets.
\newblock In {\em Proceedings of the AAAI conference on artificial
  intelligence}, volume~32, 2018.

\bibitem{bojchevski2018netgan}
Aleksandar Bojchevski, Oleksandr Shchur, Daniel Z{\"u}gner, and Stephan
  G{\"u}nnemann.
\newblock Netgan: Generating graphs via random walks.
\newblock In {\em International Conference on Machine Learning}, pages
  610--619. PMLR, 2018.

\bibitem{shi2020graphaf}
Chence Shi, Minkai Xu, Zhaocheng Zhu, Weinan Zhang, Ming Zhang, and Jian Tang.
\newblock Graphaf: a flow-based autoregressive model for molecular graph
  generation.
\newblock {\em arXiv preprint arXiv:2001.09382}, 2020.

\bibitem{liu2021graphebm}
Meng Liu, Keqiang Yan, Bora Oztekin, and Shuiwang Ji.
\newblock Graphebm: Molecular graph generation with energy-based models.
\newblock {\em arXiv preprint arXiv:2102.00546}, 2021.

\bibitem{luo2021graphdf}
Youzhi Luo, Keqiang Yan, and Shuiwang Ji.
\newblock Graphdf: A discrete flow model for molecular graph generation.
\newblock {\em arXiv preprint arXiv:2102.01189}, 2021.

\bibitem{jin2018junction}
Wengong Jin, Regina Barzilay, and Tommi Jaakkola.
\newblock Junction tree variational autoencoder for molecular graph generation.
\newblock In {\em International Conference on Machine Learning}, pages
  2323--2332. PMLR, 2018.

\bibitem{you2018graphrnn}
Jiaxuan You, Rex Ying, Xiang Ren, William Hamilton, and Jure Leskovec.
\newblock Graphrnn: Generating realistic graphs with deep auto-regressive
  models.
\newblock In {\em International Conference on Machine Learning}, pages
  5708--5717. PMLR, 2018.

\bibitem{bardes2021vicreg}
Adrien Bardes, Jean Ponce, and Yann LeCun.
\newblock Vicreg: Variance-invariance-covariance regularization for
  self-supervised learning.
\newblock {\em arXiv preprint arXiv:2105.04906}, 2021.

\bibitem{chen2020simple}
Ting Chen, Simon Kornblith, Mohammad Norouzi, and Geoffrey Hinton.
\newblock A simple framework for contrastive learning of visual
  representations.
\newblock In {\em International conference on machine learning}, pages
  1597--1607. PMLR, 2020.

\bibitem{tian2020makes}
Yonglong Tian, Chen Sun, Ben Poole, Dilip Krishnan, Cordelia Schmid, and
  Phillip Isola.
\newblock What makes for good views for contrastive learning.
\newblock {\em arXiv preprint arXiv:2005.10243}, 2020.

\bibitem{tishby2000information}
Naftali Tishby, Fernando~C Pereira, and William Bialek.
\newblock The information bottleneck method.
\newblock {\em arXiv preprint physics/0004057}, 2000.

\bibitem{alemi2016deep}
Alexander~A Alemi, Ian Fischer, Joshua~V Dillon, and Kevin Murphy.
\newblock Deep variational information bottleneck.
\newblock {\em arXiv preprint arXiv:1612.00410}, 2016.

\bibitem{Morris+2020}
Christopher Morris, Nils~M. Kriege, Franka Bause, Kristian Kersting, Petra
  Mutzel, and Marion Neumann.
\newblock Tudataset: A collection of benchmark datasets for learning with
  graphs.
\newblock In {\em ICML 2020 Workshop on Graph Representation Learning and
  Beyond (GRL+ 2020)}, 2020.

\bibitem{hu2020open}
Weihua Hu, Matthias Fey, Marinka Zitnik, Yuxiao Dong, Hongyu Ren, Bowen Liu,
  Michele Catasta, and Jure Leskovec.
\newblock Open graph benchmark: Datasets for machine learning on graphs.
\newblock {\em arXiv preprint arXiv:2005.00687}, 2020.

\bibitem{stochastic_function}
Eric~W Weisstein.
\newblock Stochastic function.
\newblock {\em MathWorld--A Wolfram Web Resource.
  https://mathworld.wolfram.com/StochasticFunction.html}.

\bibitem{giannone2015prior}
Domenico Giannone, Michele Lenza, and Giorgio~E Primiceri.
\newblock Prior selection for vector autoregressions.
\newblock {\em Review of Economics and Statistics}, 97(2):436--451, 2015.

\bibitem{kass1996selection}
Robert~E Kass and Larry Wasserman.
\newblock The selection of prior distributions by formal rules.
\newblock {\em Journal of the American statistical Association},
  91(435):1343--1370, 1996.

\bibitem{wang2019towards}
Jingkang Wang, Tianyun Zhang, Sijia Liu, Pin-Yu Chen, Jiacen Xu, Makan Fardad,
  and Bo~Li.
\newblock Towards a unified min-max framework for adversarial exploration and
  robustness.
\newblock {\em arXiv preprint arXiv:1906.03563}, 2019.

\bibitem{boyd2004convex}
Stephen Boyd, Stephen~P Boyd, and Lieven Vandenberghe.
\newblock {\em Convex optimization}.
\newblock Cambridge university press, 2004.

\bibitem{donsker1975asymptotic}
Monroe~D Donsker and SR~Srinivasa Varadhan.
\newblock Asymptotic evaluation of certain markov process expectations for
  large time, i.
\newblock {\em Communications on Pure and Applied Mathematics}, 28(1):1--47,
  1975.

\bibitem{belghazi2018mutual}
Mohamed~Ishmael Belghazi, Aristide Baratin, Sai Rajeshwar, Sherjil Ozair,
  Yoshua Bengio, Aaron Courville, and Devon Hjelm.
\newblock Mutual information neural estimation.
\newblock In {\em International Conference on Machine Learning}, pages
  531--540. PMLR, 2018.

\bibitem{wu2020graph}
Tailin Wu, Hongyu Ren, Pan Li, and Jure Leskovec.
\newblock Graph information bottleneck.
\newblock {\em arXiv preprint arXiv:2010.12811}, 2020.

\bibitem{yu2020graph}
Junchi Yu, Tingyang Xu, Yu~Rong, Yatao Bian, Junzhou Huang, and Ran He.
\newblock Graph information bottleneck for subgraph recognition.
\newblock {\em arXiv preprint arXiv:2010.05563}, 2020.

\bibitem{thakoor2021bootstrapped}
Shantanu Thakoor, Corentin Tallec, Mohammad~Gheshlaghi Azar, R{\'e}mi Munos,
  Petar Veli{\v{c}}kovi{\'c}, and Michal Valko.
\newblock Bootstrapped representation learning on graphs.
\newblock {\em arXiv preprint arXiv:2102.06514}, 2021.

\bibitem{chen2019powerful}
Ting Chen, Song Bian, and Yizhou Sun.
\newblock Are powerful graph neural nets necessary? a dissection on graph
  classification.
\newblock {\em arXiv preprint arXiv:1905.04579}, 2019.

\bibitem{xu2018powerful}
Keyulu Xu, Weihua Hu, Jure Leskovec, and Stefanie Jegelka.
\newblock How powerful are graph neural networks?
\newblock {\em arXiv preprint arXiv:1810.00826}, 2018.

\bibitem{sterling2015zinc}
Teague Sterling and John~J Irwin.
\newblock Zinc 15--ligand discovery for everyone.
\newblock {\em Journal of chemical information and modeling},
  55(11):2324--2337, 2015.

\bibitem{wu2018moleculenet}
Zhenqin Wu, Bharath Ramsundar, Evan~N Feinberg, Joseph Gomes, Caleb Geniesse,
  Aneesh~S Pappu, Karl Leswing, and Vijay Pande.
\newblock Moleculenet: a benchmark for molecular machine learning.
\newblock {\em Chemical science}, 9(2):513--530, 2018.

\bibitem{zhang2018link}
Muhan Zhang and Yixin Chen.
\newblock Link prediction based on graph neural networks.
\newblock {\em arXiv preprint arXiv:1802.09691}, 2018.

\bibitem{lin2021prototypical}
Shuai Lin, Pan Zhou, Zi-Yuan Hu, Shuojia Wang, Ruihui Zhao, Yefeng Zheng, Liang
  Lin, Eric Xing, and Xiaodan Liang.
\newblock Prototypical graph contrastive learning.
\newblock {\em arXiv preprint arXiv:2106.09645}, 2021.

\bibitem{narayanan2017graph2vec}
Annamalai Narayanan, Mahinthan Chandramohan, Rajasekar Venkatesan, Lihui Chen,
  Yang Liu, and Shantanu Jaiswal.
\newblock graph2vec: Learning distributed representations of graphs.
\newblock {\em arXiv preprint arXiv:1707.05005}, 2017.

\end{thebibliography}

\appendix

\section*{Appendix}

\section{Experiments on Classical Small Benchmarks} \label{appendx:small_bench}
We assay our methods on classical small benchmarks of MUTAG and PTC-MR datasets with the standard setting of unsupervised learning \cite{lin2021prototypical}.
We adopt the GraphCL backbone architecture and hyper-parameters in \cite{you2020graph} for the learned priors, and compare with state-of-the-art (SOTA) approaches of Graph2Vec \cite{narayanan2017graph2vec}, InfoGraph \cite{sun2019infograph}, MVGRL \cite{hassani2020contrastive}, GCC \cite{qiu2020gcc} and GraphCL \cite{you2020graph}.

Results in Table \ref{tab:small_benchmark} reach the consistent observation as in Section \ref{sec:semi_supervised_learning} result (i), that
on small benchmarks (MUTAG and PTC-MR are even smaller than those in Table \ref{tab:semi_supervised}), learned prior performs on par
with GraphCL using tedious trial-and-error on prefabricated
augmentations, as well as other SOTA methods.

\begin{table}[!htb]
\begin{center}
 \caption{Unsupervised learning on classical small benchmarks. Reported numbers are classification accuracies (\%).}
\label{tab:small_benchmark}
\resizebox{0.32\textwidth}{!}{\begin{tabular}{c c c c c c c c c c }
    \hline
    \hline
    Methods & MUTAG & PTC-MR \\
    \hline
    \hline
    Graph2Vec & 83.2$\pm$9.3 & 60.2$\pm$6.9 \\
    InfoGraph &  89.0$\pm$1.1 & 61.7$\pm$1.7 \\
    MVGRL & 89.7$\pm$1.1 & 62.5$\pm$1.7 \\
    GCC & 86.4$\pm$0.5 & 58.4$\pm$1.2 \\
    GraphCL & 86.8$\pm$1.3 & 58.4$\pm$1.7 \\
    \hline
    LP-InfoMin & 88.69$\pm$0.84 & 62.34$\pm$0.83 \\
    LP-InfoBN & 88.68$\pm$0.41 & 61.84$\pm$0.59 \\
    LP-Info(Min\&BN) & 88.94$\pm$0.69 & 61.67$\pm$0.49 \\
    \hline
    \hline
\end{tabular}}
\end{center}
\end{table}

\section{Ablation on Base Model} \label{appendx:base_model}
We validate the efficacy of learned priors on different base models.
Except for GraphCL, we choose another simple and effective variant, BRGL \cite{thakoor2021bootstrapped} for the experiment.
We examine on the semi-supervised learning setting as in Section \ref{sec:semi_supervised_learning} with the COLLAB dataset.

Results in Table \ref{tab:bgrl_backbone} show a similar phenomenon as in Section \ref{sec:experiment}, that compared with prefabricated augmentations, the learned prior leads to the better generalization especially guided by a proper principle.
We demonstrate the above observation holds with different base models.

\begin{table}[!htb]
\begin{center}
 \caption{Semi-supervised learning on COLLAB of learned priors with different base models.}
\label{tab:bgrl_backbone}
\resizebox{0.35\textwidth}{!}{\begin{tabular}{c c c c c c c c c c }
    \hline
    \hline
    Methods & GraphCL & BGRL \\
    \hline
    \hline
    None & 74.23$\pm$0.21 & 74.41$\pm$0.19 \\
    LP-InfoMin & 74.66$\pm$0.14 & 74.69$\pm$0.30 \\
    LP-InfoBN & 74.61$\pm$0.28 & 74.44$\pm$0.12 \\
    LP-Info(Min\&BN) & 74.84$\pm$0.31 & 74.87$\pm$0.12 \\
    \hline
    \hline
\end{tabular}}
\end{center}
\end{table}

\section{Ablations on Hyper-parameter of learned prior} \label{appendx:ablation}
We perform ablation studies on three key hyper-parameters of learned priors: $\gamma$ (in the last paragraph of Section \ref{sec:principle}) controlling the trade-off between InfoMin and InfoBN principles;
$\delta$ (in equation \eqref{eq:graphcl_learned_prior_reward}) depicting the value of attenuated rewards;
and threshold (in Section \ref{sec:principle}) determining when to trigger the attenuated rewards.
Experiments are conducted on the semi-supervised learning setting as in Section \ref{sec:semi_supervised_learning} with the COLLAB dataset.

Results in Table \ref{tab:ablation_gamma} state the performance of the incorporated principle is stable insensitive to the trade-off factor $\gamma$ within a certain wide range (in COLLAB around 0.3 to 0.7). Since it is dataset-dependent, we determine its value via validation per dataset.
Results in Table \ref{tab:ablation_delta} show the attenuated reward value should not be over small or large (where the positive reward is 1), and we by default set it as 0.1 which is safe.
Results in Table \ref{tab:ablation_threshold} illuminate the threshold greater than or equal to the (principled) loss mean is necessary to reach a good performance, and we by default configurate it as mean.

\begin{table}[!htb]
\begin{center}
 \caption{Ablation study on $\gamma$ of learned priors in semi-supervised learning on COLLAB.}
\label{tab:ablation_gamma}
\resizebox{0.5\textwidth}{!}{\begin{tabular}{c c c c c c c c c c }
    \hline
    \hline
    $\gamma$ & 0 & 0.1 & 0.3 & 0.5 & 0.7 & 0.9 & 1 \\
    \hline
    \hline
    Acc.(\%) & 74.66 & 74.81 & 74.86 & 74.84 & 74.87 & 74.59 & 74.61  \\
    \hline
    \hline
\end{tabular}}
\end{center}
\end{table}

\begin{table}[!htb]
\begin{center}
 \caption{Ablation study on $\delta$ of learned priors in semi-supervised learning on COLLAB.}
\label{tab:ablation_delta}
\resizebox{0.25\textwidth}{!}{\begin{tabular}{c c c c c c c c c c }
    \hline
    \hline
    $\delta$ & 0.01 & 0.1 & 0.5 \\
    \hline
    \hline
    Acc.(\%) & 74.44 & 74.84 & 74.64 \\
    \hline
    \hline
\end{tabular}}
\end{center}
\end{table}


\begin{table}[!htb]
\begin{center}
 \caption{Ablation study on the threshold value of learned priors in semi-supervised learning on COLLAB.}
\label{tab:ablation_threshold}
\resizebox{0.5\textwidth}{!}{\begin{tabular}{c c c c c c c c c c }
    \hline
    \hline
    Thres. & mean-2std & mean-std & mean & mean+std & mean+2std \\
    \hline
    \hline
    Acc.(\%) & 74.43 & 74.42 & 74.84 & 74.87 & 74.78 \\
    \hline
    \hline
\end{tabular}}
\end{center}
\end{table}

\begin{table}[!htb]
\begin{center}
 \caption{Efficiency comparison of pre-training 100 epochs in semi-supervised learning on COLLAB.}
\label{tab:efficiency}
\resizebox{0.32\textwidth}{!}{\begin{tabular}{c c c c c c c c c c }
    \hline
    \hline
    Methods & GraphCL & Generator & LP \\
    \hline
    \hline
    Time(min) & 23 & 51 & 68 \\
    \hline
    \hline
\end{tabular}}
\end{center}
\end{table}

\section{Efficiency Analysis} \label{appendx:efficiency}
Since the learned prior is composed of graph contrastive learning and graph generative model, its pre-training complexity is equal to the addition of the complexities of two components. We empirically show the running time in Table \ref{tab:efficiency}.
Although we observe that the generator is more expensive to train, its scalability on large datasets is promising as reported in \cite{kipf2016variational,wang2018graphgan,bojchevski2018netgan}.

\end{document}